\pdfoutput=1
\documentclass{article}

    \PassOptionsToPackage{numbers, compress}{natbib}

\usepackage[main, final]{neurips_2026}

\usepackage[utf8]{inputenc}
\usepackage[T1]{fontenc}
\usepackage{url}
\usepackage{booktabs}
\usepackage{amsfonts}
\usepackage{nicefrac}
\usepackage{microtype}
\usepackage{xcolor}
\usepackage[utf8]{inputenc}
\usepackage[T1]{fontenc}
\usepackage{microtype}
\usepackage{graphicx}
\usepackage{caption}
\usepackage{subcaption}
\usepackage{booktabs}
\usepackage{amsmath,amssymb,mathtools}
\usepackage{amsfonts}
\usepackage{bm}
\usepackage{xcolor}
\usepackage{siunitx}
\usepackage{enumitem}
\usepackage[bottom]{footmisc}
\usepackage{algorithm}
\usepackage{algorithmic}
\usepackage{multirow}
\usepackage{array}
\usepackage{ragged2e}
\usepackage{etoolbox}
\usepackage{wrapfig}
\usepackage{csquotes}

\newcommand{\R}{\mathbb{R}}

\newcommand{\norm}[1]{\left\lVert #1 \right\rVert}
\newcommand{\T}{^\mathsf{T}}

\newcommand{\K}{\mathbf{K}}

\newcommand{\W}{\mathbf{W}}

\newcommand{\Enc}{E_\theta}
\newcommand{\Dec}{D_\phi}
\newcommand{\shrink}{\operatorname{shrink}}
\newcommand{\blkdiag}{\operatorname{blkdiag}}
\newcommand{\argmin}{\operatorname*{arg\,min}}

\usepackage{amsthm}
\usepackage{mathrsfs}

\theoremstyle{definition}

\usepackage[colorlinks=true,allcolors=blue]{hyperref}

\usepackage{cleveref}

\crefname{section}{Section}{Sections}
\crefname{table}{Table}{Tables}
\crefname{figure}{Figure}{Figures}
\crefname{algorithm}{Algorithm}{Algorithms}
\crefname{appendix}{Appendix}{Appendices}

\setlist[itemize]{leftmargin=*}
\setlist[enumerate]{leftmargin=*}

\usepackage{placeins}
\usepackage{fontawesome5}
\hypersetup{pdftitle={Sparse Koopman Autoencoders Identify Local Dynamical Regimes in Multibasin Systems},
  pdfauthor={Aidan Li, Uday Kiran Reddy Tadipatri, Mahan Fathi, Sarath Chandar, Ross Goroshin}}
\providecommand{\crval}[1]{\csname crv@#1\endcsname}
\expandafter\def\csname crv@dy/t1/lista/h100\endcsname{${\mathbf{7.59{\times}10^{-5}}}\,{}^{\ast\ast}$}
\expandafter\def\csname crv@dy/t1/lista_bd/h100\endcsname{${8.64{\times}10^{-5}}\,{}^{\ast\ast}$}
\expandafter\def\csname crv@dy/t1/lista_sb/h100\endcsname{${8.47{\times}10^{-5}}\,{}^{\ast\ast}$}
\expandafter\def\csname crv@dy/t1/sparse_mlp_bd/h100\endcsname{${2.15{\times}10^{-4}}\,{}^{\ast\ast}$}
\expandafter\def\csname crv@dy/t1/sparse_mlp/h100\endcsname{${2.25{\times}10^{-4}}\,{}^{\ast}$}
\expandafter\def\csname crv@dy/t1/dense_mlp/h100\endcsname{$0.00617$}
\expandafter\def\csname crv@dy/t1/dmd/h100\endcsname{$0.747$}
\expandafter\def\csname crv@dy/t1/edmd_direct/h100\endcsname{$0.308$}
\expandafter\def\csname crv@dy/t1/edmd_relifted/h100\endcsname{$0.00164$}
\expandafter\def\csname crv@dy/t1/rbf_edmd_direct/h100\endcsname{$0.0194$}
\expandafter\def\csname crv@dy/t1/rbf_edmd_relifted/h100\endcsname{$0.00171$}
\expandafter\def\csname crv@dy/t1/lista/h2000\endcsname{${\mathbf{0.339}}^{\ast\ast}$}
\expandafter\def\csname crv@dy/t1/lista_bd/h2000\endcsname{${0.341}^{\ast\ast}$}
\expandafter\def\csname crv@dy/t1/lista_sb/h2000\endcsname{${0.359}^{\ast\ast}$}
\expandafter\def\csname crv@dy/t1/sparse_mlp_bd/h2000\endcsname{${0.467}^{\ast\ast}$}
\expandafter\def\csname crv@dy/t1/sparse_mlp/h2000\endcsname{${0.526}^{\ast\ast}$}
\expandafter\def\csname crv@dy/t1/dense_mlp/h2000\endcsname{$0.851$}
\expandafter\def\csname crv@dy/t1/dmd/h2000\endcsname{$3.01$}
\expandafter\def\csname crv@dy/t1/edmd_direct/h2000\endcsname{$2.40$}
\expandafter\def\csname crv@dy/t1/edmd_relifted/h2000\endcsname{$0.964$}
\expandafter\def\csname crv@dy/t1/rbf_edmd_direct/h2000\endcsname{$1.73$}
\expandafter\def\csname crv@dy/t1/rbf_edmd_relifted/h2000\endcsname{$1.00$}
\expandafter\def\csname crv@dy/t1/lista/h4000\endcsname{${\mathbf{1.27}}^{\ast\ast}$}
\expandafter\def\csname crv@dy/t1/lista_bd/h4000\endcsname{${1.29}^{\ast\ast}$}
\expandafter\def\csname crv@dy/t1/lista_sb/h4000\endcsname{${1.30}^{\ast\ast}$}
\expandafter\def\csname crv@dy/t1/sparse_mlp_bd/h4000\endcsname{${1.53}^{\ast\ast}$}
\expandafter\def\csname crv@dy/t1/sparse_mlp/h4000\endcsname{${1.60}^{\ast\ast}$}
\expandafter\def\csname crv@dy/t1/dense_mlp/h4000\endcsname{$1.97$}
\expandafter\def\csname crv@dy/t1/dmd/h4000\endcsname{$3.21$}
\expandafter\def\csname crv@dy/t1/edmd_direct/h4000\endcsname{$2.83$}
\expandafter\def\csname crv@dy/t1/edmd_relifted/h4000\endcsname{$1.85$}
\expandafter\def\csname crv@dy/t1/rbf_edmd_direct/h4000\endcsname{$2.20$}
\expandafter\def\csname crv@dy/t1/rbf_edmd_relifted/h4000\endcsname{$2.00$}

\providecommand{\crval}[1]{\csname crv@#1\endcsname}
\expandafter\def\csname crv@mb/t1/lista/h100\endcsname{${0.0292}^{\ast\ast}$}
\expandafter\def\csname crv@mb/t1/lista_bd/h100\endcsname{${0.0359}^{\ast\ast}$}
\expandafter\def\csname crv@mb/t1/lista_sb/h100\endcsname{${0.0348}^{\ast\ast}$}
\expandafter\def\csname crv@mb/t1/sparse_mlp_bd/h100\endcsname{${0.0283}^{\ast\ast}$}
\expandafter\def\csname crv@mb/t1/sparse_mlp/h100\endcsname{${\mathbf{0.0269}}^{\ast\ast}$}
\expandafter\def\csname crv@mb/t1/dense_mlp/h100\endcsname{$0.654$}
\expandafter\def\csname crv@mb/t1/dmd/h100\endcsname{$1.77$}
\expandafter\def\csname crv@mb/t1/edmd_direct/h100\endcsname{$1.91$}
\expandafter\def\csname crv@mb/t1/edmd_relifted/h100\endcsname{$1.73$}
\expandafter\def\csname crv@mb/t1/rbf_edmd_direct/h100\endcsname{$1.94$}
\expandafter\def\csname crv@mb/t1/rbf_edmd_relifted/h100\endcsname{$1.58$}
\expandafter\def\csname crv@mb/t1/lista/h500\endcsname{${0.0895}^{\ast\ast}$}
\expandafter\def\csname crv@mb/t1/lista_bd/h500\endcsname{${0.109}^{\ast\ast}$}
\expandafter\def\csname crv@mb/t1/lista_sb/h500\endcsname{${0.102}^{\ast\ast}$}
\expandafter\def\csname crv@mb/t1/sparse_mlp_bd/h500\endcsname{${0.0881}^{\ast\ast}$}
\expandafter\def\csname crv@mb/t1/sparse_mlp/h500\endcsname{${\mathbf{0.0784}}^{\ast\ast}$}
\expandafter\def\csname crv@mb/t1/dense_mlp/h500\endcsname{$2.54$}
\expandafter\def\csname crv@mb/t1/dmd/h500\endcsname{$2.81$}
\expandafter\def\csname crv@mb/t1/edmd_direct/h500\endcsname{$2.90$}
\expandafter\def\csname crv@mb/t1/edmd_relifted/h500\endcsname{$2.71$}
\expandafter\def\csname crv@mb/t1/rbf_edmd_direct/h500\endcsname{$2.86$}
\expandafter\def\csname crv@mb/t1/rbf_edmd_relifted/h500\endcsname{$2.52$}
\expandafter\def\csname crv@mb/t1/lista/h1000\endcsname{${0.106}^{\ast\ast}$}
\expandafter\def\csname crv@mb/t1/lista_bd/h1000\endcsname{${0.125}^{\ast\ast}$}
\expandafter\def\csname crv@mb/t1/lista_sb/h1000\endcsname{${0.119}^{\ast\ast}$}
\expandafter\def\csname crv@mb/t1/sparse_mlp_bd/h1000\endcsname{${0.104}^{\ast\ast}$}
\expandafter\def\csname crv@mb/t1/sparse_mlp/h1000\endcsname{${\mathbf{0.0897}}^{\ast\ast}$}
\expandafter\def\csname crv@mb/t1/dense_mlp/h1000\endcsname{$2.91$}
\expandafter\def\csname crv@mb/t1/dmd/h1000\endcsname{$2.97$}
\expandafter\def\csname crv@mb/t1/edmd_direct/h1000\endcsname{$3.03$}
\expandafter\def\csname crv@mb/t1/edmd_relifted/h1000\endcsname{$2.86$}
\expandafter\def\csname crv@mb/t1/rbf_edmd_direct/h1000\endcsname{$4.43$}
\expandafter\def\csname crv@mb/t1/rbf_edmd_relifted/h1000\endcsname{$2.64$}

\providecommand{\crval}[1]{\csname crv@#1\endcsname}
\expandafter\def\csname crv@mb/t1/lista/hbf\endcsname{${0.0348}^{\ast\ast}$}
\expandafter\def\csname crv@mb/t1/lista_bd/hbf\endcsname{${0.0560}^{\ast\ast}$}
\expandafter\def\csname crv@mb/t1/lista_sb/hbf\endcsname{${\mathbf{0.0326}}^{\ast\ast}$}
\expandafter\def\csname crv@mb/t1/sparse_mlp_bd/hbf\endcsname{${0.102}^{\ast\ast}$}
\expandafter\def\csname crv@mb/t1/sparse_mlp/hbf\endcsname{${0.0679}^{\ast\ast}$}
\expandafter\def\csname crv@mb/t1/dense_mlp/hbf\endcsname{$1.31$}
\expandafter\def\csname crv@mb/t1/lista/nfam\endcsname{$4.3$}
\expandafter\def\csname crv@mb/t1/lista_bd/nfam\endcsname{$4.2$}
\expandafter\def\csname crv@mb/t1/lista_sb/nfam\endcsname{$4.3$}
\expandafter\def\csname crv@mb/t1/sparse_mlp_bd/nfam\endcsname{$3.9$}
\expandafter\def\csname crv@mb/t1/sparse_mlp/nfam\endcsname{$4.0$}
\expandafter\def\csname crv@mb/t1/dense_mlp/nfam\endcsname{$1.0$}

\providecommand{\crval}[1]{\csname crv@#1\endcsname}

\providecommand{\crval}[1]{\csname crv@#1\endcsname}

\providecommand{\crval}[1]{\csname crv@#1\endcsname}

\providecommand{\crval}[1]{\csname crv@#1\endcsname}

\title{Sparse Koopman Autoencoders Identify Local Dynamical Regimes in Multibasin Systems}

\author{Aidan Li$^{1,2,3}$ \quad
  Uday Kiran Reddy Tadipatri$^{4}$ \quad
  Mahan Fathi$^{5}$ \\
  \textbf{Sarath Chandar}$^{1,2,6,7}$ \quad
  \textbf{Ross Goroshin}$^{2,3}$\\
  $^{1}$Chandar Research Lab \quad
  $^{2}$Mila -- Quebec AI Institute \quad
  $^{3}$Universit\'e de Montr\'eal \\
  $^{4}$University of Pennsylvania \quad
  $^{5}$NVIDIA \quad
  $^{6}$Polytechnique Montr\'eal \quad
  $^{7}$Canada CIFAR AI Chair \\
 \texttt{aidan.li@mila.quebec}\\\faGithub\enspace\href{https://github.com/chandar-lab/skae}{\texttt{github.com/chandar-lab/skae}}
}

\begin{document}

\maketitle

\begin{abstract}
    Koopman autoencoders (KAEs) seek a higher-dimensional latent representation in which nonlinear dynamics evolve linearly.
    However, many interesting systems have multiple basins of attraction, and both theoretical and empirical work has shown these multibasin systems cannot generally admit a single finite-dimensional global Koopman embedding under standard assumptions.
    We posit that encoders with a sparsity-inducing objective encouraging few active latent coefficients will provide latent supports as an inspectable basin-modeling principle for Koopman autoencoders.
    We use these encoders producing sparse latents in training Sparse Koopman Autoencoders (SKAEs) without basin labels or other regime annotations, and treat the learned latent supports as model-produced regime variables after training.
    Across a range of procedurally generated multibasin systems and chaotic flows, we show that SKAEs have superior forecasting performance compared to dense-latent KAEs.
    We also perform a mechanistic study that shows latent supports produced by SKAEs are both essential for the quality of the representation and useful for identifying basins on basin-interior states, whereas dense-latent KAEs collapse to an uninformative single support family.
    These results identify sparse latents and their corresponding supports as label-free, interpretable regime variables for Koopman learning in nonlinear systems with multiple local dynamical laws.
\end{abstract}

\section{Introduction}
\label{sec:introduction}

Learning representations for nonlinear dynamical systems remains a central problem in
machine learning, scientific computing, and control.
A particularly useful goal is to
embed nonlinear dynamics into coordinates in which the evolution is linear, or at least locally linear, because linear models can be combined with mature tools for prediction,
estimation, and control, including model predictive control and linear quadratic
regulation~\citep{mayne_constrained_2000,grune_nonlinear_2017,korda_linear_2018, kalman_general_1960, mamakoukas_local_2019}. \par This goal is also well motivated
theoretically. Classical dynamical-systems results provide local topological
linearizations near hyperbolic fixed points~\citep{grobman1959homeomorphism,
hartman1960lemma}, while Koopman eigenfunction constructions can yield linearizing
coordinates on basins of attraction for stable equilibria and periodic
orbits~\citep{lan_linearization_2013,kvalheim_existence_2021}. Thus, nonlinear systems
may admit linear dynamics with a specific change of coordinates.
\begin{figure}[t]\normalcolor
\centering
\includegraphics[width=0.95\linewidth]{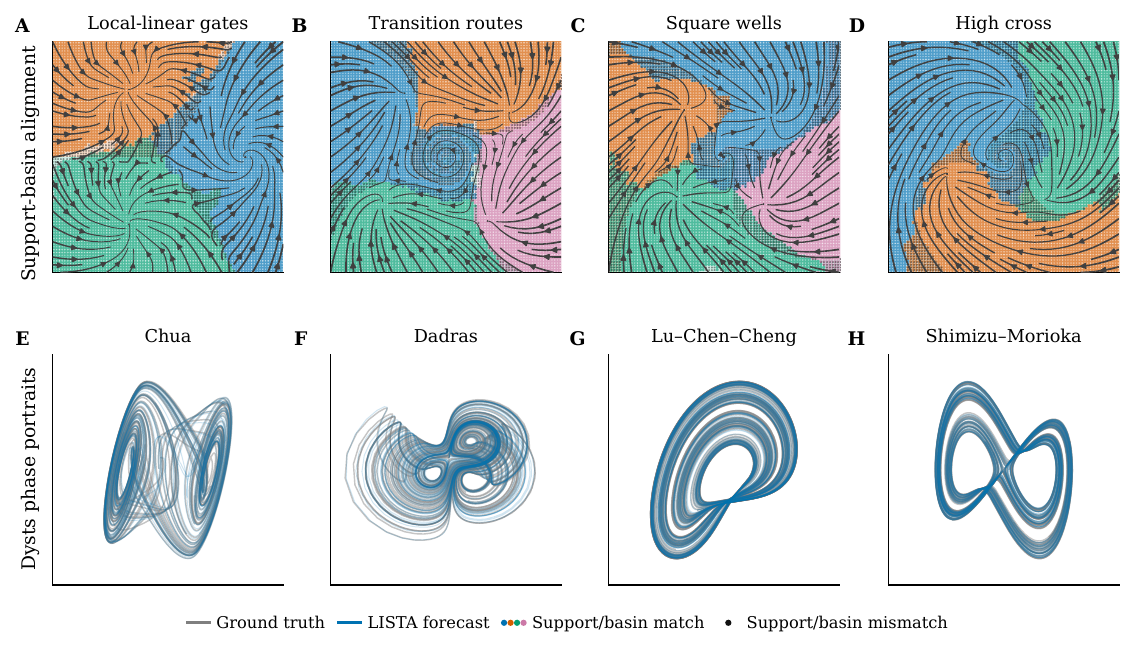}
\caption{
Top: procedurally generated multibasin systems where gray streamlines are the true vector field, colored dots mark support-family/basin matches, and black dots mark support-family/basin mismatches. The learned LISTA
(seed 0) support families are mapped post hoc to each family's dominant validation basin, which colors each dot. Bottom:  Dysts phase portraits of \(20\) test starts forecast for \(5{,}000\) observation steps, with held-out truth in gray and seed-0 LISTA forecasts in blue.}
\label{fig:benchmarks}
\end{figure}

Although these results show these useful coordinates exist, they do not provide a finite-data procedure for constructing such coordinates.
Koopman operator theory formalizes the linearization idea by lifting nonlinear
state evolution to a linear operator acting on observables \citep{koopman_hamiltonian_1931, koopman_dynamical_1932, mezic2005spectral, brunton_modern_2022}.
This viewpoint has inspired data-driven approximations of Koopman dynamics, such as DMD and EDMD \citep{schmid_dmd_2010, williams_datadriven_2015, williams_kernel-based_2016},
and more recently, Koopman autoencoders, which learn an encoder from state space to a latent representation, a linear latent transition matrix, and a decoder back to state space \citep{takeishi2017learning,lusch_deep_2018,otto2019linearly, azencot_forecasting_2020, shi_deep_2022, fathi2024course}.
These models provide an appealing route from existence results to practical learning: rather than hand-designing observables, they attempt to learn the embedding directly from trajectory data.

A central challenge for Koopman approximations is that many practically interesting nonlinear systems typically contain multiple equilibria or basins of attraction. For example, a damped mechanical system can settle into different wells, a biological or chemical model can approach different stable equilibria, and a controlled system can move between regions with distinct local dynamics.
However, theory shows that these systems cannot generally admit a single finite-dimensional global linearization with a continuous encoder-decoder pair \citep{brunton_koopman_2016, pan_lifting_2024,liu_properties_2025}, and empirical evidence supports this in practice \citep{fathi2024course}.
In such multistable systems, it is possible to linearize the dynamics within basins, but the linearizations needed in different basins of attraction are generally incompatible with each other.
Therefore, for multi-attractor systems, the right objective is not to force all basins through one continuous global linearization, but to discover and use distinct local linear regimes.

This paper proposes inducing sparse latent structure as a practical and interpretable method for Koopman learning of multi-attractor systems (\cref{fig:benchmarks}).
By inducing the state to activate only a small subset of coordinates in lifted latent space, we can incentivize different regions of the original state space to occupy different active coordinates, since nearby states should have similar latent embeddings, and thus enable local linearizations.
In this sparse representation, nonzero coefficient values in a latent vector can provide continuous coordinates for prediction within a selected region, while the active indices provide a discrete support object that can be inspected across states and used to select which columns of the latent transition matrix and the decoder act on the code\footnote{Concurrent work by \citet{kuang_lpwm_2026} explores a similar separation between latent support
and feature magnitudes in action-conditioned world models.}.

Sparse coding and LISTA-style encoders \citep{gregor_learning_2010,chen_hyperparameter_lista_2021,aberdam_ada-lista_2022} provide the right inductive bias for this purpose. They naturally induce piecewise-linear, union-of-subspaces structure~\citep{olshausen_emergence_1996,chen_atomic_1998,donoho_optimally_2003,gregor_learning_2010}, which suits this setting where different attractors or basins require different local linearizations in theory.
\par\par
We hypothesize that sparse-latent Koopman autoencoders will assign local dynamical regimes to distinct latent support families, yielding a learned regime variable without basin supervision.
Then, periodically decoding and re-encoding the state \citep{fathi2024course} can refresh the quality of the latent embedding over time by re-selecting the active support when trajectories move across the state space.\footnote{\citet{fathi2024course} introduced periodic re-encoding and also sketched an idea on sparse, block-structured latents for switching dynamics in their Appendix~D. In comparison, we examine sparsity-inducing effects with LISTA encoders and make the idea of supports as regime variables falsifiable with concrete experiments.}

\paragraph{Contributions.} We introduce Sparse Koopman Autoencoders (SKAEs), showing that sparse latent supports can serve as label-free regime variables within a single trained model with one global transition matrix. Across multibasin and chaotic benchmarks, SKAEs with either a LISTA or a sparse MLP encoder (i) improve periodically re-encoded long-horizon forecasting over dense-latent KAEs while (ii) producing supports that are functionally necessary for accurate rollouts and (iii) interpretable as emergent local dynamical regimes aligned with basin interiors, where LISTA encoders give the lowest basin-alignment entropy.

\section{Problem formulation}
\label{sec:problem}

The introduction motivates a multibasin version of Koopman learning. Given sampled trajectories of a nonlinear system, the aim is to learn a state-reconstructing latent representation whose evolution is linear, without being told which basin or local dynamical regime generated each observation.
We formulate the problem at the cadence of the stored benchmark observations.
For each system, let \(\mathcal X\subseteq\R^{d_x}\) be the observed state space and let
\begin{equation}
    x_{k+1}=F_{\Delta t}(x_k),\qquad x_k\in\mathcal X,
    \label{eq:sampled_map}
\end{equation}
where $x_k\in\R^{d_x}$ is the state at the $k$th stored sample, $F_{\Delta t}$ denotes the transformation to the next discrete observation step based on underlying dynamics, and $\Delta t$ is the system-specific observation interval.
Forecasting is therefore defined by repeatedly applying the map $F_{\Delta t}$, and a horizon of $h$ means $h$ applications of $F_{\Delta t}$. We denote $k$ compositions as $F_{\Delta t}^{\,(k)}$. Because $\Delta t$ may differ across systems, equal values of $h$ may yield different physical time.

\paragraph{Basins of attraction.} Many systems of interest are multistable: different initial conditions can approach different long-run behaviors.
A fixed point is a state $x^\ast$ satisfying $F_{\Delta t}(x^\ast)=x^\ast$.
More generally, a trajectory may approach an attractor $A$, such as a stable equilibrium, limit cycle, finite set, or chaotic invariant set.
The basin of attraction of $A$ is the set of initial conditions whose trajectories converge to $A$:
\begin{equation}
  \mathcal{B}(A)
  =
  \left\{
    x_0\in\R^{d_x}:
    \inf_{a\in A}\norm{F_{\Delta t}^{\,(k)}(x_0)-a}_2\!\to 0
    \text{ as } k\to\infty
  \right\}.
  \label{eq:basin}
\end{equation}
When multiple attractors coexist, their basins typically partition the state space up to basin boundaries. This defines the setup for this paper.
The task is to learn a representation that can support linear prediction where the appropriate linear law depends on where the state lies in the multibasin geometry.

\paragraph{Koopman theory.} The stored-step map \(F_{\Delta t}\) induces a Koopman operator \(\mathcal K\) on scalar observables \(g:\mathcal X\to\R\) by composition, \((\mathcal K g)(x)=g(F_{\Delta t}(x))\). Although \(F_{\Delta t}\) may be nonlinear, \(\mathcal K\) is linear on the space of observables. A finite-dimensional Koopman approximation searches for a vector observable \(\Phi:\mathcal X\to\R^{d_z}\) and a matrix \(K\) such that \(\Phi(F_{\Delta t}(x))\approx K\Phi(x)\), while retaining enough information to reconstruct the state.

\paragraph{Koopman autoencoders.} We study Koopman autoencoders (KAEs) in this setting. The model learns an encoder $\Enc:\R^{d_x}\rightarrow\R^{d_z}$, a decoder $\Dec:\R^{d_z}\rightarrow\R^{d_x}$, and a latent linear transition matrix $K\in\R^{d_z\times d_z}$. For observation $x_k$ at arbitrary timestep $k$, the corresponding open-loop $h$-step forecast is
\begin{equation}
  \hat x_{k+h}=\Dec\!\left(K^h\Enc(x_k)\right).
  \label{eq:kae_forecast}
\end{equation}
The same matrix $K$ is applied at every step of the rollout, so any state-dependent structure required for prediction must be encoded in the learned representation rather than in a supplied regime label.

\paragraph{Basin labels are withheld from training.} In the intended training and deployment setting, the model is not given basin labels or trajectory-to-basin assignments. When benchmark basin labels are available, they are reserved for post-hoc evaluation. They are not used to train the encoder or transition, or to select re-encoding periods.

\section{Method}
\label{sec:method}
\subsection{Sparse supports as local regime variables}
\label{sec:method_regime_supports}

A finite-dimensional Koopman representation for the multibasin systems in \cref{sec:problem} should encode two complementary quantities: (i) continuous coordinates that linearize the dynamics within a basin, and (ii) which local linear regime is active. We use sparsity to couple these two roles. If the state \(x\) is represented by a sparse latent code \(z\), then the support of \(z\) can act as a regime variable, while the nonzero coefficient values provide coordinates for the corresponding local linear representation.

Given an overcomplete dictionary \(D\), sparse coding estimates \(z\) by solving the Lasso objective
\begin{equation}
  z^\star(x)
  =
  \argmin_{z\in\R^{d_z}}
  \frac{1}{2}\norm{x-Dz}_2^2+\lambda_{\rm sc}\norm{z}_1,
  \qquad \lambda_{\rm sc}>0,
  \label{eq:sparse_coding_objective}
\end{equation}
which finds a code $z$ with few non-zero coefficients to reconstruct the input $x$. In other words, it encourages a sparse solution to the under-specified problem (classically, $\ell_0$ instead of $\ell_1$ will find the sparsest solution with the fewest non-zero coefficients).

The classical solver to this problem (ISTA \citep{beck2009fast}) can be unrolled into a depth-\(Q\) network of ISTA iterations, yielding learned ISTA (LISTA) \citep{gregor_learning_2010}. In our setting, LISTA is an encoder yielding the sparse code \(z=\Enc(x)\), where the \emph{support} is the set of active $z$ coordinates.
With a linear decoder, the support determines which decoder columns participate in reconstructing \(x\), and the associated nonzero coefficients specify coordinates within that selected reconstruction subspace. When \(D\) and \(z\) are learned jointly, as in sparse coding \citep{olshausen_emergence_1996}, this induces a data-adaptive partition of the input state space where each region can be locally reconstructed with a code having the same support.

We therefore posit that combining Koopman representations and sparse coding objectives in a sparse Koopman autoencoder (SKAE) implicitly captures multibasin attractor dynamics without explicit basin labels in training.
The Koopman objective encourages the continuous coefficients on each support to evolve linearly, while the sparsity objective encourages different local regimes to use different active coordinates.
Thus, the model can represent a multibasin system using a discrete support pattern to identify the active basin and continuous coefficients for its localized linear representation. Essentially, when using only one transition matrix \(K\) and one decoder, the support selects which of their columns act on the code.
When ground-truth basin labels are available in benchmarks, we use them only after training to assess support--basin alignment or to construct diagnostic interventions.

\subsection{Koopman autoencoder and sparse encoder}
\label{sec:method_model}

The predictor follows the KAE formulation in \cref{sec:problem}: an encoder, linear decoder, and latent transition, all of which are optimized jointly, produce the rollout in \cref{eq:kae_forecast}. As in sparse coding \citep{gregor_learning_2010}, we constrain the columns of the decoder $\Dec=D$ to be unit norm to avoid degenerate solutions. We experiment with \(K\) that is dense, block diagonal, or softly block-regularized.

\paragraph{Sparse encoding.} Our main sparse encoder is LISTA \citep{gregor_learning_2010}. It first computes a dense pre-code \(c_t=f_\theta(x_t)\), then applies learned shrinkage refinements for $q=0,\ldots,Q-1$:
\begin{equation}
  u_t^{(0)}=\shrink(c_t,\tau),\qquad
  u_t^{(q+1)}=\shrink(Su_t^{(q)}+c_t,\tau),\qquad
  z_t=u_t^{(Q)}.
  \label{eq:lista}
\end{equation}
Here, $\shrink(a,\tau)=\operatorname{sign}(a)\max(|a|-\tau,0)$
is applied element-wise, \(S\) is learned, and \(\tau\) controls the threshold scale for shrinkage. The thresholding step creates exact zeros, so the encoder explicitly selects active latent coordinates while learning the selection rule from the prediction objective. \Cref{app:experimental_details} gives the benchmark-specific output and threshold choices.

\paragraph{Transition families.}
We vary the structure of \(K\) separately from the encoder. The dense transition leaves \(K\) unrestricted. The block-diagonal transition partitions latent coordinates into \(M\) fixed groups (using the basin count of each system in the multibasin benchmark) and constrains \(K_{\rm block}=\blkdiag(K_1,\ldots,K_M)\). We also use a soft-block LISTA ablation that keeps \(K\) dense but penalizes cross-group entries, \(\mathcal{L}_{\rm offblock}=\sum_{i,j:\,g(i)\ne g(j)} |K_{ij}|\), where \(g(i)\) is the fixed architectural group containing coordinate \(i\).

\subsection{Training objective}
\label{sec:method_training}

For each training window \(x_0,\ldots,x_L\) of adjacent stored states, let \(z_t=\Enc(x_t)\) be the encoded states and \(\hat z_t=K^t z_0\) the latent rollout from the first state. Each application of \(K\) is one stored step (\cref{eq:sampled_map}). We use standard KAE losses and an \(\ell_1\) sparsity-inducing penalty on the encoded states, each averaged over the windows of a batch:
\begin{equation}
\begin{alignedat}{2}
  \mathcal{L}_{\rm pred} &= \frac{1}{L\sqrt{d_x}}\sum_{t=1}^{L}\norm{\Dec(\hat z_t)-x_t}_2, &\qquad
  \mathcal{L}_{\rm rec}  &= \frac{1}{L\sqrt{d_x}}\sum_{t=1}^{L}\norm{\Dec(z_t)-x_t}_2,\\
  \mathcal{L}_{\rm lin}  &= \frac{1}{L}\sum_{t=1}^{L}\norm{\hat z_t-z_t}_2, &\qquad
  \mathcal{L}_{\rm sp}   &= \frac{1}{L}\sum_{t=1}^{L}\norm{z_t}_1.
\end{alignedat}
\label{eq:losses}
\end{equation}
The observation-space terms are normalized by $\sqrt{d_x}$ so that their scale is comparable across systems of different state dimension. The latent terms are left in their native scale because the latent dimension is fixed within each comparison. \cref{app:experimental_details} gives the implemented loss scaling on each benchmark.
The total objective is
\begin{equation}
  \mathcal{L}
  =
  \left(
  \lambda_{\rm pred}\mathcal{L}_{\rm pred}
  +\lambda_{\rm rec}\mathcal{L}_{\rm rec}
  +\lambda_{\rm lin}\mathcal{L}_{\rm lin}
  +\lambda_{\rm sp}\mathcal{L}_{\rm sp}
  \right)
  +\mathcal{L}_{\rm struct}.
  \label{eq:training_objective}
\end{equation}
Here $\mathcal{L}_{\rm struct}=0$ for the dense and block-diagonal transition variants.
For the soft-block transition, $\mathcal{L}_{\rm struct}=\lambda_{\rm off}\mathcal{L}_{\rm offblock}$, which probes whether a soft transition grouping is sufficient to induce useful support structure.
The prediction and latent-linearity terms make the representation useful for Koopman rollout, the reconstruction term keeps the code tied to the observed state, and the sparsity term encourages selective activation.

\subsection{Support objects}
\label{sec:method_supports}

 In this subsection, we introduce several notions of support that are used at evaluation time.

 Let \(z=\Enc(x)\in\mathbb{R}^{d_z}\). A \textit{support rule} first converts the latent code $z$ into a Boolean active-coordinate vector \(m(x)  \in\{0,1\}^{d_z}\), where \(m_i(x)=1\) means that latent coordinate \(i\) is active for state \(x\). Equivalently, the exact support is the index set \(S(x)=\{i:m_i(x)=1\}\). \par Exact supports defined by an arbitrary threshold can be unstable: a small perturbation in $x$-space can lead to a different support in $z$, so a single basin may be covered by several nearby or partially overlapping active-coordinate vectors.
We therefore distinguish exact supports from \emph{support families}, which group nearby Boolean support vectors and test whether these groups form basin-scale objects. Support families in general are more robust and lead to less basin fragmentation.

We use two support constructions:
\begin{itemize}
  \item \textbf{Absolute-threshold support.} The absolute support vector is the Boolean mask denoted \(m_{{\rm abs}}(x)\), where $m_{{\rm abs},i}(x)=\mathbf{1}\{|z_i|>10^{-3}\}$ for indices $i$. \(S_{\rm abs}(x)=\{i:m_{{\rm abs},i}(x)=1\}\) is the equivalent active-index set. $m_{{\rm abs}}(x)$ is the input to the support family \(F_{\rm abs}\) (defined next).
  \item \textbf{Absolute-threshold support family.} \(F_{\rm abs}\) is a group label for exact supports, and every exact support $m_{{\rm abs}}$ is assigned an \(F_{\rm abs}\) label.
  \(F_{\rm abs}\) groups exact supports whose active-index sets substantially overlap. We define the overlap score as the Jaccard index, or intersection-over-union, between the active coordinates of two Boolean support vectors \(m_{{\rm abs}}(x)\). \par On an evaluation collection \(\mathcal X\), the algorithm first counts all distinct Boolean vectors and then visits those distinct vectors from most to least frequent, following a leader-style threshold clustering rule. Each family \(f\) stores one fixed representative vector \(r_f\): the exact support vector that created that family. For the next distinct vector \(m_{\rm abs}\), the algorithm computes the Jaccard index
  \[
    J(m_{\rm abs},r_f)=
    \frac{\sum_i \mathbf{1}\{m_{{\rm abs},i}=1\ \text{and}\ r_{f,i}=1\}}
         {\sum_i \mathbf{1}\{m_{{\rm abs},i}=1\ \text{or}\ r_{f,i}=1\}},
  \]
  with \(J(0,0)=1\) when both vectors are all zero. If the best existing representative has Jaccard similarity at least \(\tau=0.5\), every occurrence of \(m_{\rm abs}\) receives that family label. Otherwise, \(m_{\rm abs}\) starts a new family and becomes the representative \(r_f\) for that family.
  Pseudocode is provided in Appendix \ref{app:support_definitions}.
  We write \(F_{\rm abs}(x)\) for the family assigned to \(m_{\rm abs}(x)\).

\end{itemize}

\subsection{Periodic re-encoding}
If a support indicates the currently relevant local dynamics, a long rollout should be able to update that support.\label{sec:method_reencoding} We therefore evaluate periodic re-encoding, introduced by \citet{fathi2024course}. Starting from $z_k=\Enc(x_k)$ and a re-encoding period $m$, the model advances for $m$ Koopman steps, decodes the predicted state, and encodes that prediction again:
\begin{equation}
  \tilde z_{k+m}=K^m z_k,\qquad
  \tilde x_{k+m}=\Dec(\tilde z_{k+m}),\qquad
  z_{k+m}=\Enc(\tilde x_{k+m}).
  \label{eq:reencode}
\end{equation}
The same procedure is repeated over the rollout horizon. Each rollout uses only the model's predicted state, not the true future state, a basin label, or an oracle for transitions between basins. For each trained model, we tune a separate re-encoding period $m$ from a fixed grid on validation rollouts\footnote{In general, there is no universal re-encoding period that is best for all systems.}.

\section{Experiments}
\label{sec:experiments}
Our experiments aim to evaluate if finite-dimensional Koopman representations can be learned for multi-attractor systems implicitly without resorting to side information such as basin labels.
We first discuss the quality of learned Koopman representations from sparse KAEs via a forecasting-based evaluation.
Next, in the context of sparse inference, we evaluate the importance of the specific active set of coordinates for forecasting within a basin of attraction.
Finally, we evaluate alignment of support families \(F_{\rm abs}\) with withheld basin labels.

\paragraph{Benchmark systems.} The main benchmark for our experiments contains 15 procedurally generated two-dimensional multibasin systems. These systems are constructed by specifying potential wells around known attractor coordinates, and superimposing dynamics that encourage transitions to different far-away regimes.
Because we specify the locations of the potential wells, the basin labels (the attractor that a state's trajectory reaches) are well-defined and used only for evaluation. We also use a subset of 10 chaotic systems from the Dysts repository \citep{gilpin_chaos_2021,gilpin_model_2023} with about 1{,}000 stored steps per dominant period\footnote{The dominant period is given by each system's Dysts metadata: the dominant Fourier timescale, i.e.\ the period of the strongest significant peak in the power spectrum of a long trajectory.} as an additional forecasting test.
Analytic expressions, per-system timesteps, and other details are in Appendix \ref{app:multibasin_inventory}, \ref{app:dysts_inventory}.

\paragraph{Training models.} We compare six neural models shown in \cref{tab:forecasting_main} by varying the encoder and latent transition structure and studying their effects in this setting.
We test three LISTA sparse encoders (where LISTA-BD denotes block-diagonal latent transitions and LISTA-SB denotes soft-block regularization),
two sparse-latent MLPs applying the sparsity penalty $\mathcal{L}_{\rm sp}$ \eqref{eq:losses} with dense \citep{fathi2024course} or block-diagonal (BD) transitions,
and a dense-latent MLP control baseline with no sparsity incentive. All models use a linear decoder with columns having unit norm constraint.

Most of these architectures are biased toward sparse inference by varying degrees. Our contribution is in studying how different ways of inducing sparsity can be useful. For LISTA, encoded states are sparse by construction via thresholding, and encoded latents get sparsity pressure from \eqref{eq:losses}. For sparse MLPs, the encoded states have induced sparsity through the ReLU activations and $\ell_1$ regularization \citep{zhang_2016_L1,layton_2024_relu}.
The dense MLP baseline has no intended zero-producing mechanism: it uses tanh activations instead of ReLU and has no $\ell_1$ penalty.
Each model is trained for 15 random seeds on each system before evaluation. Training details are reported in Appendix \ref{app:experimental_details}.

\begin{table}[t]\normalcolor
\vspace{-1em}
\centering
\caption{Forecasting performance and basin-support diagnostics. \(H(B\mid F_{\rm abs})\) is the entropy of the basin label given the support family and \(\overline{|F_{\rm abs}|}\) is the number of support families. Values are means over systems of per-system seed IQMs. Direct baselines lift the state once whereas re-lifted ones re-lift it every \(m\) steps (all neural models use periodic re-encoding). \(\ast\ast\): lower than Dense MLP at \(p<0.05\), Holm-corrected, in every system. $\ast$ denotes the same significance in all systems but one. \textbf{Bold}: best in each \(\downarrow\) column.}
\label{tab:forecasting_main}
\label{tab:dysts_long}
{\setlength{\tabcolsep}{2pt}
\renewcommand{\arraystretch}{0.92}\resizebox{\textwidth}{!}{\begin{tabular}{@{}l ccc ccc cc@{}}
\toprule
& \multicolumn{6}{c}{Forecasting MSE\(\downarrow\)} & \multicolumn{2}{c}{Support diagnostics} \\
\cmidrule(lr){2-7}\cmidrule(l){8-9}
& \multicolumn{3}{c}{Multibasin, 15 systems} & \multicolumn{3}{c}{Dysts, 10 systems} & \multicolumn{2}{c}{Multibasin, 15 systems} \\
\cmidrule(lr){2-4}\cmidrule(lr){5-7}\cmidrule(l){8-9}
Model & H100 & H500 & H1000 & H100 & H2000 & H4000 & $H(B\!\mid\!F_{\rm abs})\downarrow$ & $\overline{|F_{\rm abs}|}$ \\
\midrule
LISTA & \crval{mb/t1/lista/h100} & \crval{mb/t1/lista/h500} & \crval{mb/t1/lista/h1000} & \crval{dy/t1/lista/h100} & \crval{dy/t1/lista/h2000} & \crval{dy/t1/lista/h4000} & \crval{mb/t1/lista/hbf} & \crval{mb/t1/lista/nfam} \\
LISTA-BD & \crval{mb/t1/lista_bd/h100} & \crval{mb/t1/lista_bd/h500} & \crval{mb/t1/lista_bd/h1000} & \crval{dy/t1/lista_bd/h100} & \crval{dy/t1/lista_bd/h2000} & \crval{dy/t1/lista_bd/h4000} & \crval{mb/t1/lista_bd/hbf} & \crval{mb/t1/lista_bd/nfam} \\
LISTA-SB & \crval{mb/t1/lista_sb/h100} & \crval{mb/t1/lista_sb/h500} & \crval{mb/t1/lista_sb/h1000} & \crval{dy/t1/lista_sb/h100} & \crval{dy/t1/lista_sb/h2000} & \crval{dy/t1/lista_sb/h4000} & \crval{mb/t1/lista_sb/hbf} & \crval{mb/t1/lista_sb/nfam} \\
Sparse MLP-BD & \crval{mb/t1/sparse_mlp_bd/h100} & \crval{mb/t1/sparse_mlp_bd/h500} & \crval{mb/t1/sparse_mlp_bd/h1000} & \crval{dy/t1/sparse_mlp_bd/h100} & \crval{dy/t1/sparse_mlp_bd/h2000} & \crval{dy/t1/sparse_mlp_bd/h4000} & \crval{mb/t1/sparse_mlp_bd/hbf} & \crval{mb/t1/sparse_mlp_bd/nfam} \\
Sparse MLP \citep{fathi2024course} & \crval{mb/t1/sparse_mlp/h100} & \crval{mb/t1/sparse_mlp/h500} & \crval{mb/t1/sparse_mlp/h1000} & \crval{dy/t1/sparse_mlp/h100} & \crval{dy/t1/sparse_mlp/h2000} & \crval{dy/t1/sparse_mlp/h4000} & \crval{mb/t1/sparse_mlp/hbf} & \crval{mb/t1/sparse_mlp/nfam} \\
\midrule
Dense MLP \emph{[baseline]} & \crval{mb/t1/dense_mlp/h100} & \crval{mb/t1/dense_mlp/h500} & \crval{mb/t1/dense_mlp/h1000} & \crval{dy/t1/dense_mlp/h100} & \crval{dy/t1/dense_mlp/h2000} & \crval{dy/t1/dense_mlp/h4000} & \crval{mb/t1/dense_mlp/hbf} & \crval{mb/t1/dense_mlp/nfam} \\
\midrule
DMD \citep{schmid_dmd_2010,tu_dynamic_2014} & \crval{mb/t1/dmd/h100} & \crval{mb/t1/dmd/h500} & \crval{mb/t1/dmd/h1000} & \crval{dy/t1/dmd/h100} & \crval{dy/t1/dmd/h2000} & \crval{dy/t1/dmd/h4000} & -- & -- \\
Poly.\ EDMD \citep{williams_datadriven_2015}, direct & \crval{mb/t1/edmd_direct/h100} & \crval{mb/t1/edmd_direct/h500} & \crval{mb/t1/edmd_direct/h1000} & \crval{dy/t1/edmd_direct/h100} & \crval{dy/t1/edmd_direct/h2000} & \crval{dy/t1/edmd_direct/h4000} & -- & -- \\
Poly.\ EDMD \citep{williams_datadriven_2015}, re-lifted & \crval{mb/t1/edmd_relifted/h100} & \crval{mb/t1/edmd_relifted/h500} & \crval{mb/t1/edmd_relifted/h1000} & \crval{dy/t1/edmd_relifted/h100} & \crval{dy/t1/edmd_relifted/h2000} & \crval{dy/t1/edmd_relifted/h4000} & -- & -- \\
RBF EDMD \citep{williams_datadriven_2015}, direct & \crval{mb/t1/rbf_edmd_direct/h100} & \crval{mb/t1/rbf_edmd_direct/h500} & \crval{mb/t1/rbf_edmd_direct/h1000} & \crval{dy/t1/rbf_edmd_direct/h100} & \crval{dy/t1/rbf_edmd_direct/h2000} & \crval{dy/t1/rbf_edmd_direct/h4000} & -- & -- \\
RBF EDMD \citep{williams_datadriven_2015}, re-lifted & \crval{mb/t1/rbf_edmd_relifted/h100} & \crval{mb/t1/rbf_edmd_relifted/h500} & \crval{mb/t1/rbf_edmd_relifted/h1000} & \crval{dy/t1/rbf_edmd_relifted/h100} & \crval{dy/t1/rbf_edmd_relifted/h2000} & \crval{dy/t1/rbf_edmd_relifted/h4000} & -- & -- \\
\bottomrule
\end{tabular}
}}
\end{table}

\begin{figure}[!t]\normalcolor
\centering
\begin{subfigure}[t]{0.49\linewidth}
  \vspace{0pt}
  \centering
  \includegraphics[width=\linewidth]{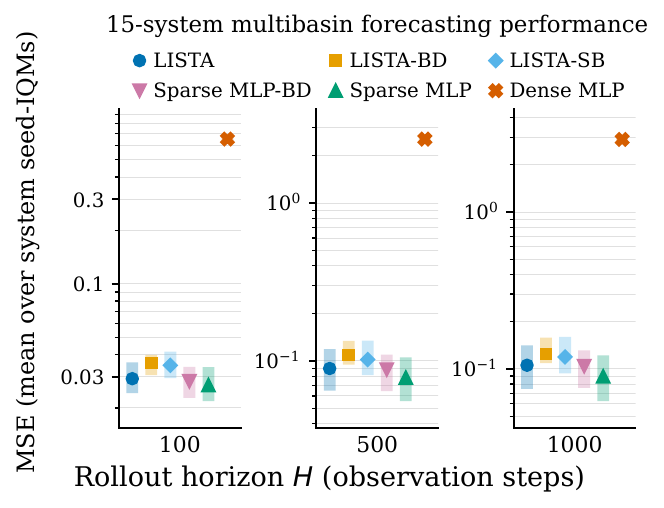}
  \caption{15-system multibasin benchmark.}
  \label{fig:fixed15_horizon_curves}
\end{subfigure}
\hfill
\begin{subfigure}[t]{0.49\linewidth}
  \vspace{0pt}
  \centering
  \includegraphics[width=\linewidth]{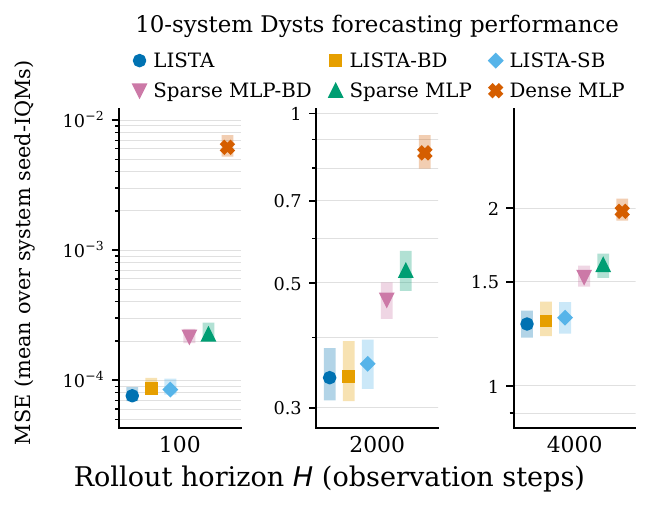}
  \caption{10-system Dysts benchmark.}
  \label{fig:dysts_long_horizon}
\end{subfigure}
\caption{Forecasting MSE at the horizons of \cref{tab:forecasting_main}, one panel per horizon. Sparse models forecast better than dense models. Markers show arithmetic means across per-system seed-IQM MSEs on a log scale. Shaded bars show fixed-system seed-bootstrap 95\% intervals after system-wise log-relative normalization, anchored to the displayed mean, so they summarize seed uncertainty within an average system.}
\label{fig:forecasting_horizon_curves}
\end{figure}

\paragraph{Statistical details.}
Point estimates are interquartile means (IQM) over seeds within each system to reduce seed uncertainty \citep{agarwal2021deep}, then arithmetic means across systems.
The markers of \cref{fig:forecasting_horizon_curves} use the same point estimate as the tables, with seed-bootstrap intervals that
summarize uncertainty over seeds for an average system.
Statistical significance is rigorously tested with one-sided paired Wilcoxon signed-rank tests and Holm corrections (Appendix \ref{app:statistical_testing}).

\subsection{Forecasting experiments}

\par

\paragraph{Sparse encoders improve forecasting.} The quality of the representation learned by a KAE is primarily measured by its ability to make accurate, multi-step forecasts.
Thus, we evaluate the forecasting performance of trained sparse and dense-latent KAEs on the two benchmarks spanning 25 systems.
Forecasting is evaluated on held-out trajectories at the stored observation cadence, with periodic re-encoding at a period chosen on validation, and summarized by mean squared error (MSE).

\begin{figure}[t]
  \centering
  \includegraphics[width=\textwidth]{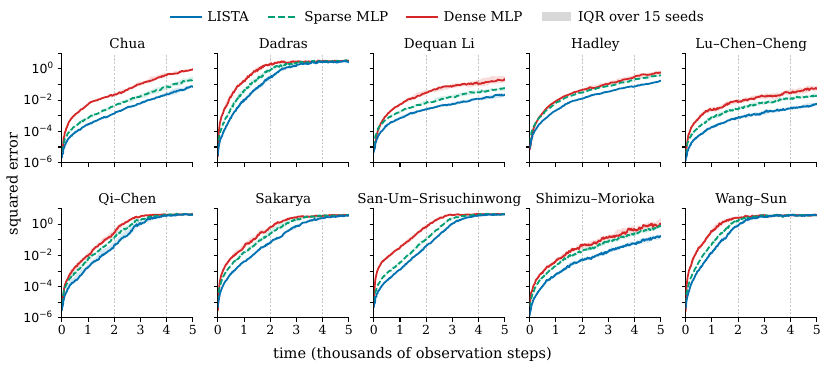}
  \caption{Forecast error over time on the ten Dysts systems, for the re-encoded forecasts of \cref{tab:forecasting_main}. At every step, we take the median squared error over the \(100\) test starts. The lines show the median over the \(15\) seeds and bands the interquartile range. Dotted lines mark the \cref{tab:forecasting_main} horizons \(H=2000\) and \(H=4000\). Each system's dominant period is about \(1{,}000\) steps. In every system, LISTA's and Sparse MLP's errors are below Dense MLP's from step \(304\) to step \(2{,}000\).}
  \label{fig:app_dysts_error_growth}
\end{figure}

The results (\cref{fig:forecasting_horizon_curves,tab:forecasting_main}) show that sparse models yield a significant reduction of MSE forecasting error at all horizons on both sets of systems. In particular, at longer horizons, the dense MLP has 23 to 32 times larger MSE at $H=1000$ on the multibasin systems and about 1.2 to 1.6 times larger MSE at $H=4000$ on the Dysts systems compared to the sparse models. \Cref{fig:app_dysts_error_growth} shows how the errors grow over time. \par On the other hand, there are marginal differences between MSE predictions among the sparse-latent encoders on the multibasin benchmark (within a factor of \(1.4\) at every horizon). These results suggest that having a sparsity-inducing encoder is crucial for good performance over all horizon prediction lengths. The appendix reports per-system results (\cref{app:per_system}), the classical baselines (\cref{app:classical_baselines}), ablations of the encoder's sparsity mechanisms (ReLU versus \(\ell_1\), \cref{app:relu_l1_ablation}) and of periodic re-encoding (\cref{app:reencoding_ablation}), and a training-window ablation (\cref{app:window_ablation}).
\par
\begin{figure}[tbp!]\normalcolor
\centering
\begin{subfigure}[t]{0.49\linewidth}\vspace{0pt}
  \centering
  \includegraphics[width=\linewidth]{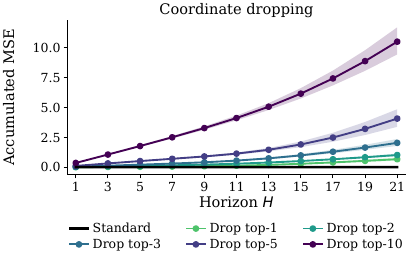}
  \caption{Dropping the largest active coordinates.}
  \label{fig:support_coordinate_dropping}
\end{subfigure}
\hfill
\begin{subfigure}[t]{0.49\linewidth}\vspace{0pt}
  \centering
  \includegraphics[width=\linewidth]{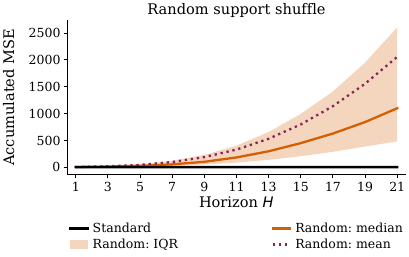}
  \caption{Moving active values to inactive coordinates.}
  \label{fig:support_coordinate_random}
\end{subfigure}
\caption{Support-coordinate interventions on a multibasin system. Interventions severely increase forecasting errors compared to the standard rollout. \textbf{(a)} The standard rollout is compared with cumulative dropping of the largest active coordinates of the initial sparse code. Shaded bands are bootstrap 95\% intervals over \(100\) initial conditions. \textbf{(b)} Active coefficient values are preserved but reassigned to randomly chosen inactive coordinates. The band is the interquartile range over shuffles and initial conditions (\(20\) shuffles each).}
\label{fig:support_coordinate_interventions}
\end{figure}

\paragraph{Sparse supports are essential for good forecasting.} If a support formed by sparse encoders is meant to carry basin-relevant dynamical information, then changing only the initial active set should damage forecasts, even when the coefficient values are otherwise kept fixed.
We do an ablation study to show whether accurate forecasts within a basin of attraction require the correct coordinates to be active by isolating the selection of sparse coordinates from their values. We do this by forcing a sparse LISTA-based KAE to use an incorrect active set with specific interventions (\cref{fig:support_coordinate_interventions}), and evaluating 21-step forecasting performance under these settings. The interventions are the following:

\begin{enumerate}
    \item \textbf{Coordinate dropping:} Force the coordinates of $z$ with the $k\in\{1,2,3,5,10\}$ largest magnitude coefficients to zero before forecasting.
    \item \textbf{Random support:} The active coefficient values are preserved, but reassigned to randomly chosen inactive latent coordinates.
\end{enumerate}

\begin{table}[tbp]
  \centering
  \centering
  \caption{Accumulated MSE at \(H=21\) for the support interventions (lower is better).}
  \label{tab:support_coordinate_h21}
  {\setlength{\tabcolsep}{2pt}
\resizebox{0.55\textwidth}{!}{\begin{tabular}{@{}lcc@{}}
\toprule
Rollout & Mean \(\pm\) SD & Median [IQR] \\
\midrule
Standard & $\mathbf{0.0112{\pm}0.0140}$ & $\mathbf{0.00735\,[0.00488,0.0122]}$ \\
Drop top-1 & $0.670{\pm}0.639$ & $0.542\,[0.189,0.700]$ \\
Drop top-2 & $1.02{\pm}0.625$ & $0.922\,[0.605,1.19]$ \\
Drop top-3 & $2.04{\pm}1.49$ & $1.67\,[1.30,2.02]$ \\
Drop top-5 & $4.07{\pm}3.82$ & $2.33\,[1.77,4.48]$ \\
Drop top-10 & $10.5{\pm}5.67$ & $8.41\,[6.22,13.3]$ \\
Random support & $2060{\pm}2590$ & $1100\,[482,2610]$ \\
\bottomrule
\end{tabular}
}}
\end{table}
\begin{figure}[tbp!]\normalcolor
\centering
\begin{subfigure}[t]{0.49\linewidth}\vspace{0pt}
  \centering
  \includegraphics[width=\linewidth]{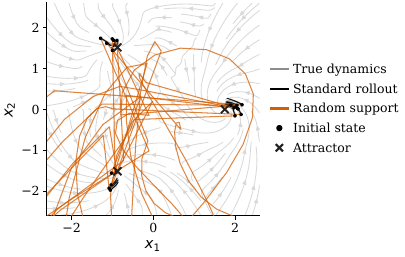}
    \caption{Trajectories after random support shuffling.}
  \label{fig:support_coordinate_trajectories}
\end{subfigure}
\hfill
\begin{subfigure}[t]{0.49\linewidth}\vspace{0pt}
  \centering
  \includegraphics[width=\linewidth]{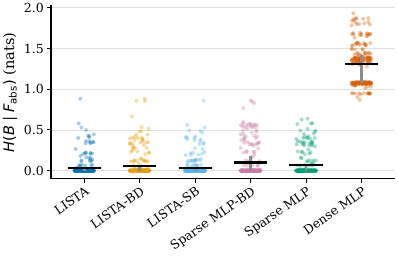}\caption{Basin-label entropy given the support family.}
    \label{fig:fixed15_entropy_strips}
\end{subfigure}
\caption{Support shuffling and support--basin alignment on the multibasin benchmark. \textbf{(a)} Rollouts after randomly shuffling the active coordinates (orange) move away from the standard rollouts (black). \textbf{(b)} Conditional entropy \(H(B\mid F_{\rm abs})\) of the basin label given the support family: the support families of the sparse models identify basins, whereas those of Dense MLP do not. Dots show system--seed pairs, gray I-bars the first-to-third quartile range of the per-system seed IQMs, and black bars the values of \cref{tab:forecasting_main}.}\label{fig:support_alignment_pair}\end{figure}

The results show that these interventions have a large detrimental effect on forecast quality, even on very short horizons. At \(H=21\), the standard rollout has mean accumulated MSE 0.0112, compared to 0.67/1.02/2.04/4.07/10.5 for the rollouts after dropping the top 1/2/3/5/10 active coordinates (\cref{tab:support_coordinate_h21}), and random support shuffling is much more destructive as seen in \cref{fig:support_coordinate_interventions}. The trajectory panel (\cref{fig:support_coordinate_trajectories}) shows the qualitative effects of the intervention.
So, supports formed by sparse encoders do not merely correlate with basin labels; the active coordinates are key to the quality of the Koopman representation.

\subsection{Support--basin alignment experiments}
\label{sec:alignment}\par

This mechanistic experiment examines the extent to which sparse support families fit after training can behave like basin-interior regime variables.
States near separatrices are challenging and may confound the inferences: near these states, support changes can be induced by small perturbations in $x$, making supports unreliable for basin identification at separatrices.
We therefore evaluate this diagnostic on held-out states deep in basins far from basin boundaries.

For a state \(x\) in basin \(b\) (the attractor that its trajectory reaches), let \(d_1(x)\) and \(d_2(x)\) be its distances to the center of \(b\) and the nearest other benchmark attractor center. The \textit{basin-depth margin} is \(d_2(x)-d_1(x)\).
This support--basin alignment experiment fits support families on validation trajectories and takes the held-out test states whose margin is positive and in the top quartile of the validation margins of their basin (the \emph{high-margin subset}, \cref{app:exp_alignment}). \par If a support family $F_{\rm abs}$ identifies a basin-specific regime used by the model, then on states deep inside a basin, it should leave little uncertainty about the withheld benchmark basin label $B$,
measured by conditional entropy \(H(B\mid F_{\rm abs})\) as the main evaluation metric. \(H(B\mid F_{\rm abs})\) should be small when support families are basin-aligned. Also, different basins should be adequately represented by distinct support families, so we count the observed number of distinct support families, denoted as \(|F_{\rm abs}|\), and compare it with the number of basins represented.

\paragraph{Support families \(F_{\rm abs}\) identify basin membership.}
As seen in \cref{tab:forecasting_main} and \cref{fig:fixed15_entropy_strips}, LISTA-based \(F_{\rm abs}\) support families have the lowest mean conditional entropy (0.0326 to 0.0560). They are followed by the sparse-latent MLPs, and then finally the dense-latent MLP baseline. \Cref{app:normalized_entropy} reports \(H(B\mid F_{\rm abs})\) normalized by the basin-label entropy \(H(B)\), which puts systems with different numbers of basins on one scale and leaves this ordering unchanged. \par LISTA-based models also expose the most support families, which better matches the benchmark's basin-count mean/median 4.20/4. Ideally, if a basin can truly be recovered by a single support family cluster, then the basin count should be one-to-one with the number of support families, $\overline{|F_{\rm abs}|}$. In contrast, the dense-latent MLP, which has no mechanism for exact zeros, collapses the representation to one support family. \par On the high-margin subset, support families fitted on validation trajectories match the basins of held-out test trajectories one-to-one in 175 of 225 system--seed cells for LISTA, 158 for Sparse MLP, and 0 for Dense MLP (\cref{app:alignment_counts}).
This suggests that, on basin-interior states, the support families induced by sparse coding may serve as a good proxy for ground-truth basin labels, and that the sparse models may have the emergent ability to discover basins unsupervised away from basin boundaries.

\par

\section{Discussion}
\label{sec:discussion}

In this paper, we show that sparse Koopman autoencoders can learn useful local regime structure in multibasin dynamical systems without observing basin labels.
Existing theory and empirical evidence indicate that multibasin systems generally should not be forced through one continuous finite-dimensional global Koopman embedding \citep{brunton_koopman_2016, pan_lifting_2024, fathi2024course, liu_properties_2025}.
Our results show that sparse codes offer a practical alternative: a single trained model \textit{can} find good representations for multiple basins by using different latent supports.
This gives practitioners a simple diagnostic to inspect after training.

For model selection, improvements in forecasting performance are more strongly driven by the sparsity incentives provided by the choice of activation functions and $\ell_1$ sparsity-inducing regularization. However, the LISTA encoder, which has an inductive bias for finding sparse and specialized representations, can still yield tangible improvements on more complicated systems (such as the chaotic systems), and it can also improve interpretability of the discovered regimes.

We qualify some limitations. First, our interpretability claims are evaluated on benchmark systems where the basin geometry is known, and the support--basin alignment analysis is intentionally restricted to states deep inside basins.
Near separatrices, small state changes can imply very different long-term outcomes, and support assignments become fragmented or unstable.
This is not a failure mode unique to sparse Koopman models, but we suggest that support families should be interpreted as reliable basin-interior regime variables rather than perfect global basin classifiers.
Moreover, support families come from a fixed post-hoc heuristic, not a learned objective. Learning them, e.g.\ contrastively, is future work. \par Second, while the experiments include a diverse set of synthetic multibasin systems and external chaotic flows, they remain deterministic and fully observed low-dimensional benchmarks, because scoring alignment needs known basins and coverage of several basin interiors.
Real-world systems may be high-dimensional, contain observation noise, partial observability, nonstationarity, control inputs, or regime changes that are not cleanly basin-like.

More broadly, this work argues that interpretability in dynamical representation learning can come from the structure of the latent code.
Sparse supports provide an interface between continuous Koopman coordinates and discrete regime structure, making them a promising tool for multi-regime forecasting, analysis, and eventually control. By showing that these supports improve prediction, are necessary for accurate rollouts, and align with withheld basin information, this paper positions sparse latent structure as a practical step toward Koopman models for multibasin dynamics.

\newpage
\begin{ack}
A.L. acknowledges support from the Fonds de recherche du Québec (FRQ) through a Master's Research Scholarship
(DOI: \url{https://doi.org/10.69777/2008193}).
S.C. is supported by the Canada CIFAR AI Chairs program, the Canada Research Chair in Lifelong Machine Learning, and the NSERC Discovery Grant.
This research was enabled in part by compute resources, software and technical help provided by Mila (\href{https://mila.quebec}{\nolinkurl{mila.quebec}}) and by the Digital Research Alliance of Canada (\href{https://alliancecan.ca}{\nolinkurl{alliancecan.ca}}).
The authors have no competing interests to declare.
\end{ack}

\bibliographystyle{plainnat}
\bibliography{biblio}

\newpage
\appendix
\crefalias{section}{appendix}\crefalias{subsection}{appendix}\FloatBarrier\section{Support families}
\label{app:support_definitions}

This appendix gives the pseudocode for the support families of \cref{sec:method_supports}.

\begin{quote}
\small
\textbf{Input:} validation states \(\mathcal X\), their masks \(m_{\rm abs}(x)\), and the Jaccard threshold \(\tau=0.5\).\\
\textbf{Output:} a family label \(F_{\rm abs}(x)\) for each \(x\in\mathcal X\), and a representative \(r_f\) for each family \(f\).
\begin{enumerate}[label=\textbf{\arabic*.},leftmargin=2.0em]
  \item Count how often each distinct mask occurs, and sort the distinct masks from most to least frequent, breaking ties by a fixed lexicographic order of the masks.
  \item For each distinct mask \(m_{\rm abs}\) in this order: if no family exists or \(\max_f J(m_{\rm abs},r_f)<\tau\), create a new family with representative \(m_{\rm abs}\); otherwise, assign \(m_{\rm abs}\) to the family that attains the maximum (the earliest one on ties).
  \item Give each state \(x\) the family of its mask \(m_{\rm abs}(x)\).
\end{enumerate}
\end{quote}

Family labels are arbitrary; only the partition they induce matters. Test states are assigned to these families as described in \cref{app:exp_alignment}.

Because the Jaccard index divides by the union of the two masks, the threshold \(\tau\) adapts to the mask size. A lower \(\tau\) merges more exact supports and can merge distinct basins; a higher \(\tau\) separates more masks and can lower \(H(B\mid F_{\rm abs})\) by fragmenting each basin. We therefore fix \(\tau=0.5\) instead of tuning it, read each entropy together with the family count, and report other thresholds in \cref{app:threshold_sensitivity}.
\par

\FloatBarrier\section{Additional experimental details}
\label{app:experimental_details}

This appendix gives the settings of the experiments in \cref{sec:experiments}; \cref{app:multibasin_inventory,app:dysts_inventory} describe the benchmarks and their data.

\subsection{Models}
\label{app:exp_models}

All neural models (\cref{tab:exp_models}) use \(d_z=256\), MLPs with two hidden layers of width \(64\), a bias-free linear decoder, and \(K\) initialized at the identity. LISTA initializes its decoder dictionary \(D_0\) from random orthonormal bases and \(S\) at \(I-D_0^\top D_0/\rho\), where \(\rho\) is \(1.05\) times a power-iteration estimate of the largest eigenvalue of \(D_0^\top D_0\), and uses \(Q=2\) refinements (\cref{eq:lista}). \Cref{tab:exp_settings} gives its benchmark-specific code and threshold.

\begin{table}[htbp]
\centering
\caption{Neural models. The last two rows are the encoder ablations of \cref{app:relu_l1_ablation}.}
\label{tab:exp_models}
{\small
\begin{tabular}{@{}l l l l@{}}
\toprule
Model & Encoder & Code sparsity & Latent transition \\
\midrule
LISTA & LISTA, ReLU-MLP pre-code & shrinkage, \(\ell_1\) & dense \\
LISTA-BD & LISTA, ReLU-MLP pre-code & shrinkage, \(\ell_1\) & block-diagonal \\
LISTA-SB & LISTA, ReLU-MLP pre-code & shrinkage, \(\ell_1\) & dense, off-block penalty \\
Sparse MLP & MLP, ReLU hidden and output & ReLU, \(\ell_1\) & dense \\
Sparse MLP-BD & MLP, ReLU hidden and output & ReLU, \(\ell_1\) & block-diagonal \\
Dense MLP & MLP, tanh hidden, linear output & none & dense \\
\midrule
Sparse MLP, \(\lambda_{\rm sp}=0\) & MLP, ReLU hidden and output & ReLU & dense \\
Dense MLP, ReLU hidden & MLP, ReLU hidden, linear output & none & dense \\
\bottomrule
\end{tabular}}
\end{table}

\subsection{Training}
\label{app:exp_training}

All neural models of a benchmark share the settings of \cref{tab:exp_settings}, except in the training-window ablation of \cref{app:window_ablation}. Every reported result uses the final weights, and no seed is selected. We fixed this recipe after exploratory sweeps over LISTA depth, shrinkage scale, sparsity coefficient, sign-split encodings, transition structure, and MLP sparsity controls, with shorter budgets or broader system lists; they are not pooled with the reported results.

\begin{table}[htbp]
\centering
\caption{Settings of the neural models. The encoder and decoder learning rate also applies to the learned LISTA thresholds on Dysts.}
\label{tab:exp_settings}
{\small
\begin{tabular}{@{}l >{\raggedright\arraybackslash}p{4.2cm} >{\raggedright\arraybackslash}p{4.2cm}@{}}
\toprule
& Multibasin & Dysts \\
\midrule
\multicolumn{3}{@{}l}{\emph{Architecture}} \\
LISTA code & \(128\) coordinates, split by sign into \(256\); decoder initialized as \([D_0,-D_0]\) & \(256\) coordinates, nonnegative shrinkage \(\max(a-\tau,0)\); decoder initialized as \(D_0\) \\
LISTA threshold \(\tau\) & fixed, \(0.15/\rho\) & learned per coordinate, initialized at \(0.15/\rho\), shared by the refinements \\
BD and SB blocks & one per basin, contiguous, near-equal size & two of \(128\) coordinates \\
\midrule
\multicolumn{3}{@{}l}{\emph{Training}} \\
Seeds & \(0,\ldots,14\) & \(0,\ldots,14\) \\
Optimization steps & \(200{,}000\) & \(12{,}000\) \\
Minibatch & \(256\) windows & \(512\) windows \\
Window length & \(L=8\) (\(9\) states) & \(L=63\) (\(64\) states) \\
Training windows & \(2{,}048\) per system and seed, fixed and cycled as \(8\) minibatches & resampled uniformly every step from the \(200\times29{,}938\) start positions of the training split \\
Model inputs & raw states & re-standardized per coordinate with training-split statistics \\
Optimizer & AdamW & AdamW \\
Learning rate, encoder and decoder & \(5{\times}10^{-5}\) & \(10^{-3}\) \\
Learning rate, \(K\) & \(5{\times}10^{-6}\) & \(2{\times}10^{-4}\) \\
Learning-rate schedule & constant & constant for \(75\%\) of the steps, then cosine decay to \(1\%\) \\
Weight decay & \(10^{-4}\), all parameters but \(K\) & none \\
\(\lambda_{\rm pred},\ \lambda_{\rm rec},\ \lambda_{\rm lin}\) & \(1,\ 0.03,\ 1\) & \(1,\ 1,\ 1\) \\
\(\lambda_{\rm sp}\) (Dense MLP: \(0\)) & \(3{\times}10^{-3}\) & \(0.03\) \\
\(\lambda_{\rm off}\) (SB models) & \(10^{-4}\) & \(10^{-4}\) \\
Gradient clipping & none & none \\
\midrule
\multicolumn{3}{@{}l}{\emph{Evaluation}} \\
Test starts per system and seed & \(100\) uniform draws in the box, separate seed stream & \(100\) windows of the test split \\
Validation starts & \(100\) uniform draws in the box, separate seed stream & \(100\) windows of the validation split \\
Horizons \(H\) (observation steps) &\(100\), \(500\), \(1000\) & \(100\), \(2000\), \(4000\) \\
Re-encoding periods \(m\) & \(10\), \(25\), \(50\), \(100\) & \(10\), \(25\), \(50\), \(100\), \(150\), \(200\) \\
\bottomrule
\end{tabular}}
\end{table}

\paragraph{Loss scaling.} On the multibasin benchmark, the four terms of \cref{eq:training_objective}, but not \(\mathcal{L}_{\rm struct}\), carry a further factor \(1/L\), so the off-block penalty weighs \(L\lambda_{\rm off}=8{\times}10^{-4}\) relative to them. On Dysts, the terms are exactly those of \cref{eq:losses}, computed on the standardized states.

\paragraph{Multibasin training windows.} Each window starts, with probability \(0.5\), from a uniform draw in the initial-condition box, and otherwise from a state of a label-free pool plus Gaussian jitter of standard deviation \(0.01\) times the box width. The pool, built before training to expose basin boundaries and transients, keeps the \(1{,}024\) top-scoring of \(4{,}096\) uniform draws, each probed for \(32\) steps and scored by its sensitivity to four Gaussian perturbations of standard deviation \(0.04\) times the box width (their mean final distance from the unperturbed state, divided by \(0.04\) times the box diagonal) plus \(0.5\) times its motion over the last \(8\) probe steps, divided by the box diagonal.

\subsection{Forecast evaluation}
\label{app:exp_forecasting}

The MSE at horizon \(H\) is the squared state error, summed over coordinates and averaged over steps \(1,\ldots,H\) and over starts; on Dysts, predictions are mapped back to the stored coordinates first. The re-encoding period (\cref{sec:method_reencoding}) of each trained model minimizes the validation MSE at \(H=500\); if no period gives a finite error, the smallest is used. If a forecast becomes non-finite or its norm exceeds \(10\) times the largest state norm in the training data, it is replaced by the mean training state from that step on.

\subsection{Support alignment and interventions}
\label{app:exp_alignment}

\paragraph{Basin labels.} For each system and seed, we draw \(64\) validation and \(64\) test trajectories of \(129\) stored states from independent uniform draws in the initial-condition box. We continue each trajectory for \(5{,}000\) steps of the true system and take the mean of its final \(500\) states as its limit point. We cluster the validation limit points by \(k\)-means into the benchmark's basin count (farthest-first initialization, at most \(25\) iterations; on every system, no two centers are closer than \(5\%\) of the box diagonal), and each state takes the label of the center nearest to its trajectory's limit point. The basin-depth margins of \cref{sec:alignment} are measured to these centers, and ties at the high-margin threshold are kept.

\paragraph{Support families.} A test support joins the family, fitted on the validation states, whose representative is most similar if their Jaccard similarity is at least \(0.5\), and is otherwise unmatched. Unmatched states count as one more family in the entropies but not in the family count. Entropies are in nats.

\paragraph{Support-coordinate interventions.} The interventions of \cref{fig:support_coordinate_interventions} use one trained LISTA model (Local-linear gates, seed \(0\)) and \(100\) starts: basin-balanced high-margin states that stay in their basin for the next \(21\) steps, taken from \(512\) separate trajectories of \(65\) stored states, which also give the basin centers and margins. An intervention edits only \(z_0=\Enc(x_0)\), whose active set is \(S_{\rm abs}(x_0)\) (\cref{sec:method_supports}), and the rollout is not re-encoded. The accumulated MSE at horizon \(H\) sums over steps \(1,\ldots,H\) the per-step squared error averaged over state coordinates, whereas the forecast MSE sums over coordinates and averages over steps.

\subsection{Compute}
\label{app:exp_compute}

The neural models were trained on NVIDIA L40S or H100 GPUs and the classical baselines on CPUs, for about \(90\) GPU hours and \(200\) CPU hours in total. Per-step cost differs between architectures, so equal step counts do not imply equal floating-point operations.
\par

\FloatBarrier\section{Statistical testing protocol}
\label{app:statistical_testing}

The IQM of \cref{sec:experiments} is the mean of the seed values between the \(25\)th and \(75\)th percentiles, inclusive, with linearly interpolated percentiles; for \(15\) distinct values, it averages the middle \(7\).

\paragraph{Paired test.} Significance marks summarize one-sided paired Wilcoxon signed-rank tests, run within each system over training seeds, with the candidate and the baseline paired by system and seed; we never pool seeds and systems into one sample. For MSE, the paired effect of system \(s\), seed \(u\), and horizon \(h\) is
\begin{equation*}
  \Delta_{s,u,h}=\log_{10} E^{\rm cand}_{s,u,h}-\log_{10} E^{\rm base}_{s,u,h},
\end{equation*}
with the alternative \(\Delta<0\), where \(E\) is the test MSE of the forecast in \cref{tab:forecasting_main} (re-encoded for the neural models; the row's rollout for the classical baselines). Log ratios match the multiplicative question and reduce the influence of heavy-tailed seed-level MSEs. For \(H(B\mid F_{\rm abs})\), the effect is the raw difference against Dense MLP, with the same alternative. The family count is not tested, because closeness to the basin count is not monotone in either direction. Every reported Wilcoxon test has all \(15\) paired seeds and no zero differences.

\paragraph{Holm correction.} The per-system \(p\)-values of one candidate, baseline, metric, and horizon form one family, which is Holm-corrected at \(\alpha=0.05\): if \(p_{(1)}\le\cdots\le p_{(m)}\) are the raw \(p\)-values, the adjusted value at rank \(i\) is \(\max_{j\le i}\min\{1,(m-j+1)\,p_{(j)}\}\). Every comparison is also tested in the reverse direction, as its own family. In \cref{tab:forecasting_main,tab:sparsity_mechanisms}, \(\ast\ast\) means that the corrected test passes in every system, and \(\ast\) that it passes in all but one; in the per-system tables of \cref{app:per_system}, a mark means that the corrected test passes in that system.

\paragraph{Comparisons.} The same test compares the sparse models with Dense MLP (\cref{tab:forecasting_main}) and with each classical baseline and rollout (\cref{app:classical_baselines}). In the encoder ablation (\cref{app:relu_l1_ablation}), LISTA, Sparse MLP, and its \(\ell_1\)-free variant are each compared with the ReLU-hidden Dense MLP; the ReLU-hidden and \(\ell_1\)-free encoders with Dense MLP; and Sparse MLP with its \(\ell_1\)-free variant. On Dysts, all models of one seed, neural and classical, are evaluated on the same \(100\) test starts.

\paragraph{Interventions.} The intervals of \cref{fig:support_coordinate_interventions} describe the variability of one model on one system and are not hypothesis tests; the coordinate-dropping bands use \(2{,}000\) bootstrap resamples of the starts.

\paragraph{Bootstrap intervals.} The \(95\%\) intervals of \cref{fig:forecasting_horizon_curves} resample the seeds within each system \(10{,}000\) times; each interval is the \(2.5\)th to \(97.5\)th percentile of the mean over systems of the log change of the seed IQM, applied to the value of \cref{tab:forecasting_main}. The paired intervals of \cref{app:relu_l1_ablation} draw one resample of the seeds per system and use it for every model, \(2{,}000\) times.
\par

\FloatBarrier\section{Additional forecasting results}
\label{app:forecasting_results}

\subsection{Classical Koopman baselines}
\label{app:classical_baselines}

We compare the sparse KAEs with three classical Koopman approximations. Each standardizes the state with the training mean and standard deviation, lifts it by a fixed dictionary, fits one linear map between consecutive lifted states by ridge regression, and reads the state back from the degree-one features:
\begin{itemize}[itemsep=1pt,parsep=0pt,topsep=2pt]
\item \textbf{DMD} \citep{schmid_dmd_2010,tu_dynamic_2014}: the state itself; the map is linear on the multibasin benchmark and affine, \(x_{t+1}=Ax_t+b\), on Dysts.
\item \textbf{Polynomial EDMD} \citep{williams_datadriven_2015}: all monomials of the state up to degree three, including the constant (\(10\) features in two dimensions and \(20\) in three).
\item \textbf{RBF EDMD} \citep{williams_datadriven_2015}: the constant, the state, and \(N\) Gaussian features \(\exp(-\|x-c\|^2/(2\sigma^2))\). The run seed draws the centers \(c\) from the standardized training states, and \(\sigma\) is the median distance over \(8{,}192\) random pairs of a \(1{,}024\)-state subsample of them.
\end{itemize}
Direct rollouts lift the initial state once; re-lifted rollouts decode and lift the state again every \(m\) steps, the analog of periodic re-encoding (for DMD, the two coincide). Test starts, horizons, and error are those of the neural models (\cref{app:exp_forecasting}); \cref{tab:classical_grids} gives the training data and selection.

\begin{table}[htbp]
\centering
\caption{Training data and selection of the classical baselines. The ridge and, for re-lifted rollouts, the period \(m\) are selected jointly, as is \(N\) for RBF EDMD. The Dysts grids were fixed before the selection on the ten systems; the last row counts the selections at an edge of the grid.}
\label{tab:classical_grids}
{\small
\begin{tabular}{@{}l >{\raggedright\arraybackslash}p{5.15cm} >{\raggedright\arraybackslash}p{5.15cm}@{}}
\toprule
& Multibasin & Dysts \\
\midrule
Training data & the \(2{,}048\) windows of the neural run with the same seed & the full training split \\
Fits & one per system and seed & DMD and polynomial EDMD: one per system, so their seeds replicate only the test starts; RBF EDMD: one per system and seed \\
Selection starts & validation starts of the same seed & validation starts of a reserved seed (\(9000\)) that enters no reported result, per system \\
Selection criterion & lowest validation MSE at \(H=500\) of the row's own rollout, non-finite last; ties to fewer features, then the larger ridge, then the shorter period & lowest finite validation MSE at \(H=500\) of the row's own rollout \\
Ridge \(\lambda\) & \(0\), \(10^{-10}\), \(10^{-8}\), \(10^{-6}\), \(10^{-4}\), \(10^{-2}\), \(10^{-1}\), \(1\), \(10\); added to \(\Phi^\top\Phi/n\) & \(10^{-12}\), \(10^{-10}\), \ldots, \(10^{-2}\); applied as \(\lambda\,\mathrm{tr}(G)/p\) for the Gram matrix \(G\) of the \(p\) features \\
Period \(m\) (re-lifted) & \(10\), \(25\), \(50\), \(100\) & \(10\), \(25\), \(50\), \(100\), \(150\), \(200\) \\
RBF features \(N\) & \(256\), \(1024\) & \(256\), \(1024\) \\
\midrule
Edge selections & RBF EDMD: \(N=1024\) in \(120\) of \(225\) direct and \(98\) of \(225\) re-lifted fits; \(\lambda=0\) in \(41\) re-lifted fits (four systems) and no direct fit; \(m=10\) in \(59\) and \(m=100\) in \(89\) re-lifted fits; polynomial EDMD: \(\lambda=0\) in \(3\) re-lifted fits (two systems), \(m=10\) in \(45\) and \(m=100\) in \(64\) re-lifted fits; DMD: \(\lambda=0\) in \(1\) fit & \(\lambda=10^{-12}\): re-lifted polynomial and RBF EDMD on \(7\) of \(10\) systems each, direct RBF EDMD on \(3\); \(\lambda=10^{-2}\): DMD on \(8\), direct polynomial EDMD on \(1\) (Sakarya); \(m=10\) in \(18\) of \(20\) re-lifted choices; \(N=1024\) on every system \\
\bottomrule
\end{tabular}
}
\end{table}

\Cref{tab:forecasting_main} gives the mean errors of the classical baselines and \cref{app:per_system} the per-system values; the tests are those of \cref{app:statistical_testing}. On the multibasin benchmark, every sparse model has significantly lower error than every classical baseline, with either rollout, in all \(15\) systems at every horizon.

Among the multibasin baselines, DMD and polynomial EDMD are about as accurate as Dense MLP at \(H=1000\), and re-lifted RBF EDMD is the strongest at every horizon. The error of direct RBF EDMD grows after the selection horizon, because selection at \(H=500\) does not penalize slightly unstable fitted maps (spectral radius up to \(1.007\)).

On Dysts, the outcome depends on the system, and the sparse KAEs and re-lifted EDMD win on different systems:
\begin{itemize}[itemsep=1pt,parsep=0pt,topsep=2pt]
\item \textbf{DMD.} LISTA and Sparse MLP have significantly lower error in all \(10\) systems at \(H=100\) and \(H=2000\).
\item \textbf{Direct EDMD.} LISTA and Sparse MLP have significantly lower error than both direct rollouts in all \(10\) systems at \(H=2000\). At \(H=100\), LISTA beats direct polynomial EDMD in \(9\) systems (not Hadley) and Sparse MLP in \(8\); direct polynomial EDMD is significantly lower than Sparse MLP and Sparse MLP-BD in Hadley, and direct RBF EDMD than every sparse model in Dequan Li, Hadley, Lu--Chen--Cheng, and Shimizu--Morioka.
\item \textbf{DMD and direct EDMD at \(H=4000\).} All three are significantly lower than every sparse model in Wang--Sun and than Sparse MLP in Dadras (direct RBF EDMD also than Sparse MLP-BD), and LISTA is significantly lower than each of them in the other \(9\) systems.
\item \textbf{Re-lifted EDMD.} Re-lifted polynomial and RBF EDMD are significantly lower than every sparse model in Dequan Li, Hadley, Lu--Chen--Cheng, Qi--Chen, and Shimizu--Morioka at every horizon, often by orders of magnitude, and significantly higher than LISTA in the other five.
\end{itemize}

\subsection{Sparsity mechanisms of the encoder}
\label{app:relu_l1_ablation}

The MLP encoders of \cref{tab:exp_models} form a ladder in which each step changes one component: ReLU hidden layers (the ReLU-hidden Dense MLP, whose code is still dense), a ReLU on the code (Sparse MLP with \(\lambda_{\rm sp}=0\)), and the \(\ell_1\) penalty (Sparse MLP). The two intermediate encoders are trained like the main models (\cref{app:experimental_details}); runs with the same seed share their training windows and validation and test starts, and Dense MLP and the two intermediate encoders also share their initial weights.

On this benchmark, the forecasting gain over Dense MLP comes from ReLU encoders, mostly from their hidden layers (\cref{tab:sparsity_mechanisms}):
\begin{itemize}[itemsep=1pt,parsep=0pt,topsep=2pt]
\item \textbf{ReLU hidden layers} make about half of the hidden activations exactly zero and lower the error of Dense MLP in every system. On the log scale, this step closes \(79\%\) to \(85\%\) of the gap in mean MSE between Dense MLP and the \(\ell_1\)-free Sparse MLP.
\item \textbf{A ReLU on the code} closes the rest: the \(\ell_1\)-free Sparse MLP has a \(1.56\), \(1.97\), and \(1.97\) times lower mean MSE than the ReLU-hidden Dense MLP at \(H=100\), \(500\), and \(1000\) (paired seed-bootstrap \(95\%\) intervals \([1.34,1.89]\), \([1.37,2.82]\), and \([1.36,2.85]\); median per-system ratio \(2.6\) to \(2.8\)), significantly in most systems. No encoder with a sparse code has significantly higher error than the ReLU-hidden Dense MLP in any system.
\item \textbf{The \(\ell_1\) penalty} makes the code sparser in all \(225\) paired runs, but it lowers the error significantly in only one system (SNIC multi-attractor) at each horizon, and raises it significantly in none.
\end{itemize}

\begin{table}[tbp]
\centering
\caption{Sparsity mechanisms of the encoder on the multibasin benchmark (\(15\) systems, \(15\) paired seeds; tests of \cref{app:statistical_testing}). MSE: means over systems of per-system seed IQMs; \(\ast\ast\): significantly lower than Dense MLP in all \(15\) systems; \(\S[1/15]\): significantly lower than Sparse MLP with \(\lambda_{\rm sp}=0\) in \(1\) system at every horizon. Sparsity rows, as means over systems on \(64\) held-out trajectories per run: exactly zero encoded coordinates; coordinates with \(|z_i|\le10^{-3}\) in the direct \(K^h\) rollout over \(8\) steps; and exactly zero activations in each hidden layer of the encoder MLP (for LISTA, its pre-code MLP).}
\label{tab:sparsity_mechanisms}
{\small
\begin{tabular}{@{}l ccccc@{}}
\toprule
& Dense MLP & Dense MLP, & Sparse MLP, & & \\
& (tanh hidden) & ReLU hidden & \(\lambda_{\rm sp}=0\) & Sparse MLP\(^{\S[1/15]}\) & LISTA \\
\midrule
\multicolumn{6}{@{}l}{\emph{Forecasting MSE}\(\downarrow\)} \\
\quad \(H=100\) & $0.654$ & ${0.0541}^{\ast\ast}$ & ${0.0347}^{\ast\ast}$ & ${0.0269}^{\ast\ast}$ & ${0.0292}^{\ast\ast}$ \\
\quad \(H=500\) & $2.54$ & ${0.196}^{\ast\ast}$ & ${0.0997}^{\ast\ast}$ & ${0.0784}^{\ast\ast}$ & ${0.0895}^{\ast\ast}$ \\
\quad \(H=1000\) & $2.91$ & ${0.229}^{\ast\ast}$ & ${0.116}^{\ast\ast}$ & ${0.0897}^{\ast\ast}$ & ${0.106}^{\ast\ast}$ \\
\multicolumn{6}{@{}l}{\emph{Systems with significantly lower / higher MSE than the ReLU-hidden Dense MLP} (\(H=100/500/1000\))} \\
\quad lower & 0/0/0 & -- & 12/10/11 & 14/14/14 & 13/13/13 \\
\quad higher & 15/15/15 & -- & 0/0/0 & 0/0/0 & 0/0/0 \\
\multicolumn{6}{@{}l}{\emph{Fraction of inactive code coordinates}} \\
\quad encoded & $1.11{\times}10^{-7}$ & $2.73{\times}10^{-8}$ & $0.524$ & $0.634$ & $0.607$ \\
\quad rollout & $0.0140$ & $0.0114$ & $0.518$ & $0.622$ & $0.615$ \\
\multicolumn{6}{@{}l}{\emph{Fraction of exactly zero hidden activations}} \\
\quad layer 1 & $8.41{\times}10^{-9}$ & $0.498$ & $0.499$ & $0.499$ & $0.492$ \\
\quad layer 2 & $0$ & $0.538$ & $0.491$ & $0.497$ & $0.559$ \\
\bottomrule
\end{tabular}
}
\end{table}

\subsection{Re-encoding ablation}
\label{app:reencoding_ablation}

Periodic re-encoding (\cref{sec:method_reencoding}) passes the predicted state back through the encoder, which corrects the latent coefficients and lets the active set change. Each segment between re-encodings is linear in the latent space, but the full forecaster, including the forecasts of \cref{tab:forecasting_main}, is an autonomous nonlinear composite of decoder, encoder, and \(K\), not a pure \(K^h\) rollout. We separate the two effects with three rollouts of each trained multibasin model:
\begin{enumerate}[label=(\alph*),itemsep=1pt,parsep=0pt,topsep=2pt]
\item no re-encoding: the direct \(K^h\) rollout;
\item support-constrained re-encoding: re-encode every \(m\) steps, but keep only the coordinates in the initial active set \(S_{\rm abs}(x_0)\);
\item unrestricted re-encoding.
\end{enumerate}
Conditions (b) and (c) share one period per model and system, chosen from the grid of \cref{tab:exp_settings} by the mean over seeds of the log validation MSE of (c) at \(H=500\); test starts never enter the choice. Condition (c) thus differs from the reported forecasts, whose period is chosen per trained model, only in sharing its period across seeds. Dense MLP codes have essentially no exact zeros, and nearly every coordinate clears \(|z_i|>10^{-3}\), so the constraint in (b) removes few Dense MLP coordinates, and (b) stays much closer to (c) than for the sparse models.

Re-encoding helps even when the active set cannot change, so part of its gain is a generic re-projection; letting the support change adds a much larger gain for the sparse models than for Dense MLP (\cref{tab:reencoding_ablation}). Both findings hold when (b) is run at its own validation-selected period.

\begin{table}[tbp]
\centering
\caption{Re-encoding ablation on the multibasin benchmark at \(H=100\): (a) no re-encoding, (b) support-constrained, and (c) unrestricted re-encoding. Cells: system--seed cells in which the first condition has strictly lower test MSE than the second (there are no ties). Ratio: geometric mean over the \(225\) cells of the test-MSE ratio of (c) to (b). Own period: (b) at the period that minimizes its own validation error.}
\label{tab:reencoding_ablation}
{\small
\begin{tabular}{@{}l ccc c cc@{}}
\toprule
& \multicolumn{3}{c}{Cells with lower test MSE} & Ratio & \multicolumn{2}{c}{(b) at its own period} \\
\cmidrule(lr){2-4}\cmidrule(lr){5-5}\cmidrule(l){6-7}
Model & (b) \(<\) (a) & (c) \(<\) (b) & (c) \(<\) (a) & (c)/(b) & (b) \(<\) (a) & (c) \(<\) (b) \\
\midrule
LISTA & 204/225 & 224/225 & 225/225 & $0.022$ & 208/225 & 224/225 \\
Sparse MLP & 205/225 & 225/225 & 225/225 & $0.015$ & 208/225 & 225/225 \\
Dense MLP & 213/225 & 192/225 & 225/225 & $0.72$ & 223/225 & 175/225 \\
\bottomrule
\end{tabular}
}
\end{table}

The validation-selected period of the reported forecasts is the smallest grid value, \(m=10\), in \(1{,}486\) of the \(1{,}575\) multibasin runs of the six main models and of Sparse MLP with \(\lambda_{\rm sp}=0\), at the edge of the grid; most Dysts fits select \(m=50\) (\(662\) of \(900\)), and none selects \(m=200\).

\subsection{Training-window ablation}
\label{app:window_ablation}

Every multibasin training window applies \(K\) only \(L=8\) times, while the direct rollouts apply it up to \(1{,}000\) times. To test whether the short window limits the direct rollouts, we retrain LISTA, Sparse MLP, and Dense MLP with seeds \(0,\ldots,4\) on windows of \(64\) states (\(L=63\)) and change nothing else. In detail:
\begin{itemize}[itemsep=1pt,parsep=0pt,topsep=2pt]
\item the factor \(1/L\) of the multibasin loss rescales all four terms equally;
\item seed \(n\) shares its initial weights, window start states, and validation and test starts with the main run of seed \(n\).
\end{itemize}

Longer windows make the sparse direct rollouts more accurate than DMD and polynomial EDMD (fit on the \(9\)-state windows, as in \cref{tab:forecasting_main}) up to \(H=500\), but periodic re-encoding remains necessary for the reported accuracy at \(H=500\) and \(H=1000\) (\cref{tab:window_ablation}):
\begin{itemize}[itemsep=1pt,parsep=0pt,topsep=2pt]
\item \textbf{Direct rollouts.} The sparse direct test MSE is lower than with \(9\)-state windows in all \(15\) systems at \(H=100\) and \(H=500\) and in \(13\) at \(H=1000\) (sign test \(p\le0.004\)).
\item \textbf{Direct rollouts against the baselines.} The sparse direct rollouts are lower than DMD and both polynomial EDMD rollouts in all \(15\) systems at \(H=100\) and in \(14\) at \(H=500\), but higher than DMD in \(12\) and \(11\) systems at \(H=1000\). They are lower than the Dense MLP direct rollout in all \(15\) systems at \(H=100\), in \(11\) (LISTA) and \(14\) (Sparse MLP) at \(H=500\), and in \(13\) at \(H=1000\).
\item \textbf{Gain from re-encoding.} With \(64\)-state windows, re-encoding still lowers the error in all \(15\) systems at \(H=500\) and \(H=1000\). At \(H=100\), it does so in only \(12\) and \(10\) systems (sign test \(p=0.02\) and \(0.15\)), partly because many of these models select long periods (\(m\ge50\) in \(29\) of the \(75\) fits of each sparse model) and so re-encode at most once by \(H=100\).
\item \textbf{Re-encoded forecasts of the sparse models.} They do not change consistently: they are lower in \(6\) to \(9\) of the \(15\) systems at each horizon (sign test \(p\ge0.3\)), with per-system ratios from \(0.03\) to \(48\); the largest increase is Sparse MLP on Arrested spiral.
\item \textbf{Re-encoded forecasts of Dense MLP.} They improve in \(11\), \(14\), and \(14\) systems (\(p=0.06\), \(5{\times}10^{-4}\), and \(5{\times}10^{-4}\)), so its gap to the sparse models at \(H=1000\) narrows from about \(29\)--\(31\)-fold to \(18\)--\(19\)-fold; both sparse models still have lower re-encoded error than Dense MLP in all \(15\) systems at every horizon.
\end{itemize}

\begin{table}[tbp]
\centering
\caption{Training-window ablation on the multibasin benchmark (\(15\) systems, seeds \(0,\ldots,4\), paired with the main runs of the same seeds, so the \(9\)-state rows differ from the \(15\)-seed values of \cref{tab:forecasting_main}). Test MSE as means over systems of per-system seed IQMs, for the direct rollout and the reported re-encoded forecast; lower is better.}
\label{tab:window_ablation}
{\small
\begin{tabular}{@{}ll ccc ccc@{}}
\toprule
& & \multicolumn{3}{c}{Direct rollout} & \multicolumn{3}{c}{Re-encoded (reported)} \\
\cmidrule(lr){3-5}\cmidrule(l){6-8}
Model & Window & H100 & H500 & H1000 & H100 & H500 & H1000 \\
\midrule
LISTA & \(9\) states & \(14.3\) & \(64.9\) & \(40.1\) & \(0.0273\) & \(0.0843\) & \(0.098\) \\
 & \(64\) states & \(0.027\) & \(2.16\) & \(25\) & \(0.0232\) & \(0.0704\) & \(0.0825\) \\
\midrule
Sparse MLP & \(9\) states & \(16.3\) & \(46.8\) & \(25.3\) & \(0.0305\) & \(0.0804\) & \(0.0916\) \\
 & \(64\) states & \(0.0234\) & \(1.23\) & \(14.4\) & \(0.0217\) & \(0.0653\) & \(0.0775\) \\
\midrule
Dense MLP & \(9\) states & \(21.6\) & \(99.7\) & \(75.1\) & \(0.665\) & \(2.53\) & \(2.82\) \\
 & \(64\) states & \(0.199\) & \(2.69\) & \(71.2\) & \(0.189\) & \(0.959\) & \(1.49\) \\
\bottomrule
\end{tabular}
}
\end{table}

\FloatBarrier\section{Additional support--basin alignment results}
\label{app:alignment_results}

This appendix extends \cref{sec:alignment}. All results are on its high-margin subset of held-out test states, with the protocol of \cref{app:exp_alignment}.

\subsection{Exact one-family-per-basin counts}
\label{app:alignment_counts}

A low \(H(B\mid F_{\rm abs})\) does not require one family per basin: it is also low when each basin splits into several families. We therefore count the system--seed cells in which families and basins correspond one-to-one (\emph{exact matches}), that is, in which
\begin{itemize}[nosep,leftmargin=1.5em]
  \item every test state joins one of the families fitted on the validation states;
  \item the family determines the basin, \(H(B\mid F_{\rm abs})=0\); and
  \item the basin determines the family, \(H(F_{\rm abs}\mid B)=0\).
\end{itemize}

Every cell contains test states of at least two basins, so a single family, as in Dense MLP, never matches (\cref{tab:alignment_counts}). The five sparse models match in every seed of nine of the \(15\) systems, and most of their misses are in Arrested spiral, Octagonal eight-well, and Triple-well Duffing.

\begin{table}[tbp]
\centering
\caption{Exact one-family-per-basin matches on the high-margin subset: seeds of each system, out of \(15\) (last row: system--seed cells, out of \(225\)), whose test support families correspond one-to-one to the basins. Kept: mean fraction of the test states in the high-margin subset, the same for every model.}
\label{tab:alignment_counts}
{\small
\setlength{\tabcolsep}{4pt}\begin{tabular}{@{}l c cccccc@{}}
\toprule
System & Kept & LISTA & LISTA-BD & LISTA-SB & \begin{tabular}[b]{@{}c@{}}Sparse\\ MLP-BD\end{tabular} & \begin{tabular}[b]{@{}c@{}}Sparse\\ MLP\end{tabular} & \begin{tabular}[b]{@{}c@{}}Dense\\ MLP\end{tabular} \\
\midrule
Local-linear gates & \(0.224\) & 15 & 15 & 15 & 15 & 15 & 0 \\
Transfer-gated local-linear & \(0.202\) & 15 & 13 & 15 & 15 & 11 & 0 \\
Arrested spiral & \(0.349\) & 2 & 1 & 0 & 0 & 0 & 0 \\
Asymmetric three-well & \(0.323\) & 15 & 15 & 15 & 15 & 15 & 0 \\
High-cross three-well & \(0.255\) & 15 & 15 & 15 & 15 & 15 & 0 \\
Hexagonal six-well & \(0.255\) & 14 & 10 & 11 & 7 & 10 & 0 \\
Octagonal eight-well & \(0.309\) & 2 & 0 & 1 & 0 & 0 & 0 \\
Pentagonal five-well & \(0.272\) & 15 & 15 & 15 & 15 & 15 & 0 \\
Square four-well & \(0.400\) & 15 & 15 & 15 & 15 & 15 & 0 \\
Triple-well Duffing & \(0.374\) & 0 & 0 & 0 & 0 & 0 & 0 \\
SNIC multi-attractor & \(0.243\) & 15 & 15 & 15 & 15 & 15 & 0 \\
Transition-routes four-well & \(0.231\) & 15 & 15 & 15 & 15 & 15 & 0 \\
Depth-gradient four-well & \(0.304\) & 15 & 15 & 15 & 15 & 15 & 0 \\
Diamond four-well & \(0.352\) & 15 & 15 & 15 & 15 & 15 & 0 \\
L-shaped five-well & \(0.217\) & 7 & 4 & 7 & 0 & 2 & 0 \\
\midrule
Total (of \(225\)) & & 175 & 163 & 169 & 157 & 158 & 0 \\
\bottomrule
\end{tabular}
}
\end{table}

\subsection{Normalized conditional entropy}
\label{app:normalized_entropy}

\(H(B\mid F_{\rm abs})\) is at most the basin-label entropy \(H(B)\), which grows with the number of basins (\(3\) to \(8\) per system). Dividing it by \(H(B)\), computed on the same states, gives a ratio in \([0,1]\) (\cref{tab:normalized_entropy}): \(0\) when the family determines the basin and \(1\) when it carries no information about it, as for the single family of Dense MLP.

\begin{table}[tbp]
\centering
\caption{Conditional entropy of the basin label given the support family on the high-margin subset, unnormalized (as in \cref{tab:forecasting_main}) and divided by the basin-label entropy \(H(B)\): mean over the \(15\) systems of per-system seed IQMs; lower is better.}
\label{tab:normalized_entropy}
{\small
\setlength{\tabcolsep}{4pt}\begin{tabular}{@{}l cccccc@{}}
\toprule
& LISTA & LISTA-BD & LISTA-SB & \begin{tabular}[b]{@{}c@{}}Sparse\\ MLP-BD\end{tabular} & \begin{tabular}[b]{@{}c@{}}Sparse\\ MLP\end{tabular} & \begin{tabular}[b]{@{}c@{}}Dense\\ MLP\end{tabular} \\
\midrule
\(H(B\mid F_{\rm abs})\) (nats) & \(0.0348\) & \(0.0560\) & \(0.0326\) & \(0.102\) & \(0.0679\) & \(1.31\) \\
\(H(B\mid F_{\rm abs})/H(B)\) & \(0.0221\) & \(0.0375\) & \(0.0206\) & \(0.0740\) & \(0.0470\) & \(1.00\) \\
\bottomrule
\end{tabular}
}
\end{table}

\subsection{Threshold sensitivity}
\label{app:threshold_sensitivity}

We recompute the alignment over absolute thresholds \(\{10^{-4},10^{-3},10^{-2}\}\) (the bound on \(|z_i|\) that defines \(S_{\rm abs}\)) and Jaccard thresholds \(\{0.3,0.4,0.5,0.6,0.7\}\) for the family merge and the assignment of test states; the paper uses \(10^{-3}\) and \(0.5\). For LISTA and Sparse MLP, at every absolute threshold (\cref{tab:threshold_sensitivity}):
\begin{itemize}[nosep,leftmargin=1.5em]
  \item Jaccard thresholds below \(0.5\) merge the supports of different basins: the entropy rises and the number of exact matches falls.
  \item Thresholds of \(0.6\) and \(0.7\) lower the entropy and raise the number of exact matches relative to \(0.5\), so our fixed choice of \(0.5\) (\cref{app:support_definitions}) is not necessarily the best one.
\end{itemize}
LISTA has a lower normalized entropy than Sparse MLP at every grid point except the absolute threshold \(10^{-2}\) with Jaccard thresholds \(0.6\) and \(0.7\), where the two are within \(0.002\); the exact-match counts favor neither model consistently. Dense MLP, omitted from the table, forms a single family at every grid point.

\begin{table}[tbp]
\centering
\caption{Threshold sensitivity of the support--basin alignment on the high-margin subset. Each cell gives the mean of \(H(B\mid F_{\rm abs})/H(B)\) over the \(225\) system--seed cells (lower is better) and, in parentheses, the number of exact matches. Bold: the paper's setting; its entropy is a mean over cells, not the mean of per-system seed IQMs of \cref{tab:normalized_entropy}.}
\label{tab:threshold_sensitivity}
{\small
\setlength{\tabcolsep}{5pt}\begin{tabular}{@{}l c ccccc@{}}
\toprule
& & \multicolumn{5}{c}{Jaccard threshold} \\
\cmidrule(l){3-7}
Model & Absolute threshold & \(0.3\) & \(0.4\) & \(0.5\) & \(0.6\) & \(0.7\) \\
\midrule
LISTA & \(10^{-4}\) & \(0.258\) (\(99\)) & \(0.107\) (\(149\)) & \(0.035\) (\(169\)) & \(0.005\) (\(194\)) & \(0.001\) (\(177\)) \\
 & \(10^{-3}\) & \(0.200\) (\(114\)) & \(0.092\) (\(151\)) & \textbf{\boldmath\(0.032\) (\(175\))} & \(0.004\) (\(195\)) & \(0.000\) (\(181\)) \\
 & \(10^{-2}\) & \(0.127\) (\(134\)) & \(0.056\) (\(156\)) & \(0.018\) (\(178\)) & \(0.006\) (\(193\)) & \(0.001\) (\(182\)) \\
\midrule
Sparse MLP & \(10^{-4}\) & \(0.327\) (\(70\)) & \(0.130\) (\(135\)) & \(0.073\) (\(153\)) & \(0.020\) (\(190\)) & \(0.005\) (\(192\)) \\
 & \(10^{-3}\) & \(0.272\) (\(91\)) & \(0.109\) (\(142\)) & \textbf{\boldmath\(0.055\) (\(158\))} & \(0.014\) (\(193\)) & \(0.002\) (\(192\)) \\
 & \(10^{-2}\) & \(0.150\) (\(132\)) & \(0.070\) (\(152\)) & \(0.024\) (\(185\)) & \(0.005\) (\(199\)) & \(0.001\) (\(189\)) \\
\bottomrule
\end{tabular}
}
\end{table}

\FloatBarrier\section{Per-system forecasting results}
\label{app:per_system}

This appendix gives the per-system test MSE behind the forecasting columns of \cref{tab:forecasting_main}, one table per benchmark and horizon. Each cell is the IQM over the \(15\) seeds of one system; the means of the columns over systems are the values in \cref{tab:forecasting_main}. The marks on the sparse models are the per-system tests of \cref{app:statistical_testing}.

\subsection{Multibasin benchmark}
\label{app:per_system_multibasin}

Every sparse-model cell in \cref{tab:persystem_mb_h100,tab:persystem_mb_h500,tab:persystem_mb_h1000} carries all three marks.

\begin{table}[tbp]
\centering
\caption{Multibasin test MSE per system at \(H=100\): IQM over \(15\) seeds; lower is better. Top: sparse models; bottom: baselines. \(\ast\), \(\dagger\), \(\ddagger\): significantly lower MSE than Dense MLP, DMD, or both polynomial EDMD rollouts in that system (Holm-corrected one-sided paired Wilcoxon test, \(p<0.05\)). Tests against RBF EDMD are not marked.}
\label{tab:persystem_mb_h100}
{\footnotesize\setlength{\tabcolsep}{3pt}\begin{tabular*}{\linewidth}{@{\extracolsep{\fill}}l ccccc@{}}
\toprule
System & LISTA & LISTA-BD & LISTA-SB & Sparse MLP-BD & Sparse MLP \\
\midrule
\addlinespace[1pt]
Local-linear gates & ${9.78{\times}10^{-4}}\,{}^{\ast\dagger\ddagger}$ & ${8.28{\times}10^{-4}}\,{}^{\ast\dagger\ddagger}$ & ${9.41{\times}10^{-4}}\,{}^{\ast\dagger\ddagger}$ & ${4.45{\times}10^{-4}}\,{}^{\ast\dagger\ddagger}$ & ${5.38{\times}10^{-4}}\,{}^{\ast\dagger\ddagger}$ \\
Transfer-gated local-linear & ${0.235}^{\ast\dagger\ddagger}$ & ${0.271}^{\ast\dagger\ddagger}$ & ${0.236}^{\ast\dagger\ddagger}$ & ${0.218}^{\ast\dagger\ddagger}$ & ${0.223}^{\ast\dagger\ddagger}$ \\
Arrested spiral & ${0.0122}^{\ast\dagger\ddagger}$ & ${0.00927}^{\ast\dagger\ddagger}$ & ${0.0147}^{\ast\dagger\ddagger}$ & ${0.0125}^{\ast\dagger\ddagger}$ & ${0.0124}^{\ast\dagger\ddagger}$ \\
Asymmetric three-well & ${0.00814}^{\ast\dagger\ddagger}$ & ${0.0266}^{\ast\dagger\ddagger}$ & ${0.0157}^{\ast\dagger\ddagger}$ & ${0.00751}^{\ast\dagger\ddagger}$ & ${0.0112}^{\ast\dagger\ddagger}$ \\
High-cross three-well & ${0.0835}^{\ast\dagger\ddagger}$ & ${0.0865}^{\ast\dagger\ddagger}$ & ${0.105}^{\ast\dagger\ddagger}$ & ${0.0709}^{\ast\dagger\ddagger}$ & ${0.0771}^{\ast\dagger\ddagger}$ \\
Hexagonal six-well & ${0.0210}^{\ast\dagger\ddagger}$ & ${0.0245}^{\ast\dagger\ddagger}$ & ${0.0293}^{\ast\dagger\ddagger}$ & ${0.0235}^{\ast\dagger\ddagger}$ & ${0.0143}^{\ast\dagger\ddagger}$ \\
Octagonal eight-well & ${0.0413}^{\ast\dagger\ddagger}$ & ${0.0646}^{\ast\dagger\ddagger}$ & ${0.0645}^{\ast\dagger\ddagger}$ & ${0.0480}^{\ast\dagger\ddagger}$ & ${0.0354}^{\ast\dagger\ddagger}$ \\
Pentagonal five-well & ${0.0110}^{\ast\dagger\ddagger}$ & ${0.0171}^{\ast\dagger\ddagger}$ & ${0.0187}^{\ast\dagger\ddagger}$ & ${0.00671}^{\ast\dagger\ddagger}$ & ${0.0114}^{\ast\dagger\ddagger}$ \\
Square four-well & ${0.00144}^{\ast\dagger\ddagger}$ & ${0.00238}^{\ast\dagger\ddagger}$ & ${0.00229}^{\ast\dagger\ddagger}$ & ${0.00282}^{\ast\dagger\ddagger}$ & ${0.00454}^{\ast\dagger\ddagger}$ \\
Triple-well Duffing & ${0.00540}^{\ast\dagger\ddagger}$ & ${0.00830}^{\ast\dagger\ddagger}$ & ${0.00494}^{\ast\dagger\ddagger}$ & ${0.00667}^{\ast\dagger\ddagger}$ & ${0.00560}^{\ast\dagger\ddagger}$ \\
SNIC multi-attractor & ${2.73{\times}10^{-4}}\,{}^{\ast\dagger\ddagger}$ & ${3.04{\times}10^{-4}}\,{}^{\ast\dagger\ddagger}$ & ${3.00{\times}10^{-4}}\,{}^{\ast\dagger\ddagger}$ & ${2.08{\times}10^{-4}}\,{}^{\ast\dagger\ddagger}$ & ${2.02{\times}10^{-4}}\,{}^{\ast\dagger\ddagger}$ \\
Transition-routes four-well & ${0.00973}^{\ast\dagger\ddagger}$ & ${0.00890}^{\ast\dagger\ddagger}$ & ${0.00874}^{\ast\dagger\ddagger}$ & ${0.0113}^{\ast\dagger\ddagger}$ & ${0.00135}^{\ast\dagger\ddagger}$ \\
Depth-gradient four-well & ${0.00155}^{\ast\dagger\ddagger}$ & ${0.00216}^{\ast\dagger\ddagger}$ & ${0.00409}^{\ast\dagger\ddagger}$ & ${0.00193}^{\ast\dagger\ddagger}$ & ${0.00124}^{\ast\dagger\ddagger}$ \\
Diamond four-well & ${0.00466}^{\ast\dagger\ddagger}$ & ${0.00662}^{\ast\dagger\ddagger}$ & ${0.00460}^{\ast\dagger\ddagger}$ & ${0.00727}^{\ast\dagger\ddagger}$ & ${0.00405}^{\ast\dagger\ddagger}$ \\
L-shaped five-well & ${0.00227}^{\ast\dagger\ddagger}$ & ${0.00905}^{\ast\dagger\ddagger}$ & ${0.0107}^{\ast\dagger\ddagger}$ & ${0.00634}^{\ast\dagger\ddagger}$ & ${0.00105}^{\ast\dagger\ddagger}$ \\
\bottomrule
\end{tabular*}

\smallskip

\begin{tabular*}{\linewidth}{@{\extracolsep{\fill}}l cccccc@{}}
\toprule
 & & & \multicolumn{2}{c}{Poly.\ EDMD} & \multicolumn{2}{c}{RBF EDMD} \\
\cmidrule(lr){4-5}\cmidrule(l){6-7}
System & Dense MLP & DMD & direct & re-lifted & direct & re-lifted \\
\midrule
Local-linear gates & $0.161$ & $1.28$ & $1.32$ & $0.714$ & $1.25$ & $0.194$ \\
Transfer-gated local-linear & $0.534$ & $1.15$ & $2.57$ & $0.821$ & $1.03$ & $0.646$ \\
Arrested spiral & $0.752$ & $1.82$ & $1.68$ & $1.66$ & $2.05$ & $1.86$ \\
Asymmetric three-well & $0.613$ & $1.65$ & $1.81$ & $1.81$ & $1.95$ & $1.56$ \\
High-cross three-well & $2.06$ & $2.37$ & $2.19$ & $2.17$ & $2.92$ & $2.26$ \\
Hexagonal six-well & $1.11$ & $1.59$ & $1.69$ & $1.69$ & $1.88$ & $1.70$ \\
Octagonal eight-well & $2.06$ & $2.73$ & $2.90$ & $2.92$ & $3.19$ & $3.02$ \\
Pentagonal five-well & $1.14$ & $1.87$ & $1.91$ & $1.91$ & $2.11$ & $2.01$ \\
Square four-well & $0.190$ & $1.96$ & $2.10$ & $2.11$ & $2.14$ & $1.81$ \\
Triple-well Duffing & $0.0540$ & $1.35$ & $0.404$ & $0.109$ & $0.480$ & $0.0464$ \\
SNIC multi-attractor & $0.0468$ & $0.465$ & $0.212$ & $0.212$ & $0.822$ & $0.0550$ \\
Transition-routes four-well & $0.200$ & $2.06$ & $2.98$ & $2.98$ & $2.24$ & $2.24$ \\
Depth-gradient four-well & $0.144$ & $2.14$ & $2.18$ & $2.19$ & $2.29$ & $2.14$ \\
Diamond four-well & $0.151$ & $2.39$ & $2.55$ & $2.56$ & $2.18$ & $2.18$ \\
L-shaped five-well & $0.586$ & $1.80$ & $2.09$ & $2.14$ & $2.63$ & $1.98$ \\
\bottomrule
\end{tabular*}
}
\end{table}

\begin{table}[tbp]
\centering
\caption{Multibasin test MSE per system at \(H=500\), as in \cref{tab:persystem_mb_h100}.}
\label{tab:persystem_mb_h500}
{\footnotesize\setlength{\tabcolsep}{3pt}\begin{tabular*}{\linewidth}{@{\extracolsep{\fill}}l ccccc@{}}
\toprule
System & LISTA & LISTA-BD & LISTA-SB & Sparse MLP-BD & Sparse MLP \\
\midrule
\addlinespace[1pt]
Local-linear gates & ${0.00154}^{\ast\dagger\ddagger}$ & ${0.00146}^{\ast\dagger\ddagger}$ & ${0.00109}^{\ast\dagger\ddagger}$ & ${5.81{\times}10^{-4}}\,{}^{\ast\dagger\ddagger}$ & ${8.28{\times}10^{-4}}\,{}^{\ast\dagger\ddagger}$ \\
Transfer-gated local-linear & ${0.745}^{\ast\dagger\ddagger}$ & ${0.862}^{\ast\dagger\ddagger}$ & ${0.751}^{\ast\dagger\ddagger}$ & ${0.664}^{\ast\dagger\ddagger}$ & ${0.662}^{\ast\dagger\ddagger}$ \\
Arrested spiral & ${0.0200}^{\ast\dagger\ddagger}$ & ${0.0168}^{\ast\dagger\ddagger}$ & ${0.0230}^{\ast\dagger\ddagger}$ & ${0.0200}^{\ast\dagger\ddagger}$ & ${0.0165}^{\ast\dagger\ddagger}$ \\
Asymmetric three-well & ${0.0304}^{\ast\dagger\ddagger}$ & ${0.0728}^{\ast\dagger\ddagger}$ & ${0.0511}^{\ast\dagger\ddagger}$ & ${0.0401}^{\ast\dagger\ddagger}$ & ${0.0403}^{\ast\dagger\ddagger}$ \\
High-cross three-well & ${0.156}^{\ast\dagger\ddagger}$ & ${0.157}^{\ast\dagger\ddagger}$ & ${0.212}^{\ast\dagger\ddagger}$ & ${0.146}^{\ast\dagger\ddagger}$ & ${0.155}^{\ast\dagger\ddagger}$ \\
Hexagonal six-well & ${0.0353}^{\ast\dagger\ddagger}$ & ${0.0395}^{\ast\dagger\ddagger}$ & ${0.0497}^{\ast\dagger\ddagger}$ & ${0.0392}^{\ast\dagger\ddagger}$ & ${0.0247}^{\ast\dagger\ddagger}$ \\
Octagonal eight-well & ${0.0612}^{\ast\dagger\ddagger}$ & ${0.0912}^{\ast\dagger\ddagger}$ & ${0.0907}^{\ast\dagger\ddagger}$ & ${0.0692}^{\ast\dagger\ddagger}$ & ${0.0514}^{\ast\dagger\ddagger}$ \\
Pentagonal five-well & ${0.0210}^{\ast\dagger\ddagger}$ & ${0.0315}^{\ast\dagger\ddagger}$ & ${0.0318}^{\ast\dagger\ddagger}$ & ${0.0110}^{\ast\dagger\ddagger}$ & ${0.0210}^{\ast\dagger\ddagger}$ \\
Square four-well & ${0.00100}^{\ast\dagger\ddagger}$ & ${0.00182}^{\ast\dagger\ddagger}$ & ${0.00182}^{\ast\dagger\ddagger}$ & ${0.00242}^{\ast\dagger\ddagger}$ & ${0.00908}^{\ast\dagger\ddagger}$ \\
Triple-well Duffing & ${0.0409}^{\ast\dagger\ddagger}$ & ${0.0453}^{\ast\dagger\ddagger}$ & ${0.0472}^{\ast\dagger\ddagger}$ & ${0.0391}^{\ast\dagger\ddagger}$ & ${0.0434}^{\ast\dagger\ddagger}$ \\
SNIC multi-attractor & ${3.79{\times}10^{-4}}\,{}^{\ast\dagger\ddagger}$ & ${3.63{\times}10^{-4}}\,{}^{\ast\dagger\ddagger}$ & ${4.73{\times}10^{-4}}\,{}^{\ast\dagger\ddagger}$ & ${2.97{\times}10^{-4}}\,{}^{\ast\dagger\ddagger}$ & ${2.60{\times}10^{-4}}\,{}^{\ast\dagger\ddagger}$ \\
Transition-routes four-well & ${0.165}^{\ast\dagger\ddagger}$ & ${0.225}^{\ast\dagger\ddagger}$ & ${0.174}^{\ast\dagger\ddagger}$ & ${0.184}^{\ast\dagger\ddagger}$ & ${0.111}^{\ast\dagger\ddagger}$ \\
Depth-gradient four-well & ${9.41{\times}10^{-4}}\,{}^{\ast\dagger\ddagger}$ & ${0.00163}^{\ast\dagger\ddagger}$ & ${0.00886}^{\ast\dagger\ddagger}$ & ${0.00141}^{\ast\dagger\ddagger}$ & ${7.27{\times}10^{-4}}\,{}^{\ast\dagger\ddagger}$ \\
Diamond four-well & ${0.0587}^{\ast\dagger\ddagger}$ & ${0.0674}^{\ast\dagger\ddagger}$ & ${0.0634}^{\ast\dagger\ddagger}$ & ${0.0887}^{\ast\dagger\ddagger}$ & ${0.0384}^{\ast\dagger\ddagger}$ \\
L-shaped five-well & ${0.00536}^{\ast\dagger\ddagger}$ & ${0.0216}^{\ast\dagger\ddagger}$ & ${0.0240}^{\ast\dagger\ddagger}$ & ${0.0156}^{\ast\dagger\ddagger}$ & ${5.81{\times}10^{-4}}\,{}^{\ast\dagger\ddagger}$ \\
\bottomrule
\end{tabular*}

\smallskip

\begin{tabular*}{\linewidth}{@{\extracolsep{\fill}}l cccccc@{}}
\toprule
 & & & \multicolumn{2}{c}{Poly.\ EDMD} & \multicolumn{2}{c}{RBF EDMD} \\
\cmidrule(lr){4-5}\cmidrule(l){6-7}
System & Dense MLP & DMD & direct & re-lifted & direct & re-lifted \\
\midrule
Local-linear gates & $0.226$ & $2.66$ & $2.53$ & $1.42$ & $2.70$ & $0.238$ \\
Transfer-gated local-linear & $1.12$ & $2.25$ & $3.68$ & $2.39$ & $2.01$ & $1.62$ \\
Arrested spiral & $2.52$ & $2.42$ & $2.37$ & $2.36$ & $2.54$ & $2.47$ \\
Asymmetric three-well & $2.81$ & $2.64$ & $2.66$ & $2.66$ & $2.68$ & $2.65$ \\
High-cross three-well & $4.77$ & $2.94$ & $2.90$ & $2.83$ & $3.05$ & $2.79$ \\
Hexagonal six-well & $4.14$ & $2.42$ & $2.43$ & $2.45$ & $2.60$ & $2.48$ \\
Octagonal eight-well & $7.47$ & $4.19$ & $4.21$ & $4.24$ & $4.40$ & $4.31$ \\
Pentagonal five-well & $4.66$ & $2.79$ & $2.79$ & $2.77$ & $2.79$ & $2.76$ \\
Square four-well & $0.282$ & $2.90$ & $2.92$ & $2.92$ & $2.92$ & $2.90$ \\
Triple-well Duffing & $0.342$ & $1.07$ & $1.04$ & $0.623$ & $1.02$ & $0.271$ \\
SNIC multi-attractor & $0.152$ & $0.769$ & $0.763$ & $0.783$ & $0.863$ & $0.175$ \\
Transition-routes four-well & $2.95$ & $4.90$ & $5.01$ & $5.01$ & $4.90$ & $4.90$ \\
Depth-gradient four-well & $0.235$ & $3.00$ & $2.99$ & $3.01$ & $2.99$ & $3.00$ \\
Diamond four-well & $0.975$ & $4.03$ & $4.04$ & $4.04$ & $4.08$ & $4.08$ \\
L-shaped five-well & $5.41$ & $3.16$ & $3.19$ & $3.17$ & $3.35$ & $3.22$ \\
\bottomrule
\end{tabular*}
}
\end{table}

\begin{table}[tbp]
\centering
\caption{Multibasin test MSE per system at \(H=1000\), as in \cref{tab:persystem_mb_h100}.}
\label{tab:persystem_mb_h1000}
{\footnotesize\setlength{\tabcolsep}{3pt}\begin{tabular*}{\linewidth}{@{\extracolsep{\fill}}l ccccc@{}}
\toprule
System & LISTA & LISTA-BD & LISTA-SB & Sparse MLP-BD & Sparse MLP \\
\midrule
\addlinespace[1pt]
Local-linear gates & ${0.00167}^{\ast\dagger\ddagger}$ & ${0.00161}^{\ast\dagger\ddagger}$ & ${0.00110}^{\ast\dagger\ddagger}$ & ${6.34{\times}10^{-4}}\,{}^{\ast\dagger\ddagger}$ & ${8.73{\times}10^{-4}}\,{}^{\ast\dagger\ddagger}$ \\
Transfer-gated local-linear & ${0.811}^{\ast\dagger\ddagger}$ & ${0.913}^{\ast\dagger\ddagger}$ & ${0.824}^{\ast\dagger\ddagger}$ & ${0.724}^{\ast\dagger\ddagger}$ & ${0.711}^{\ast\dagger\ddagger}$ \\
Arrested spiral & ${0.0210}^{\ast\dagger\ddagger}$ & ${0.0175}^{\ast\dagger\ddagger}$ & ${0.0241}^{\ast\dagger\ddagger}$ & ${0.0209}^{\ast\dagger\ddagger}$ & ${0.0170}^{\ast\dagger\ddagger}$ \\
Asymmetric three-well & ${0.0332}^{\ast\dagger\ddagger}$ & ${0.0785}^{\ast\dagger\ddagger}$ & ${0.0555}^{\ast\dagger\ddagger}$ & ${0.0440}^{\ast\dagger\ddagger}$ & ${0.0440}^{\ast\dagger\ddagger}$ \\
High-cross three-well & ${0.166}^{\ast\dagger\ddagger}$ & ${0.166}^{\ast\dagger\ddagger}$ & ${0.225}^{\ast\dagger\ddagger}$ & ${0.155}^{\ast\dagger\ddagger}$ & ${0.165}^{\ast\dagger\ddagger}$ \\
Hexagonal six-well & ${0.0371}^{\ast\dagger\ddagger}$ & ${0.0413}^{\ast\dagger\ddagger}$ & ${0.0522}^{\ast\dagger\ddagger}$ & ${0.0411}^{\ast\dagger\ddagger}$ & ${0.0260}^{\ast\dagger\ddagger}$ \\
Octagonal eight-well & ${0.0642}^{\ast\dagger\ddagger}$ & ${0.0943}^{\ast\dagger\ddagger}$ & ${0.0941}^{\ast\dagger\ddagger}$ & ${0.0718}^{\ast\dagger\ddagger}$ & ${0.0534}^{\ast\dagger\ddagger}$ \\
Pentagonal five-well & ${0.0223}^{\ast\dagger\ddagger}$ & ${0.0332}^{\ast\dagger\ddagger}$ & ${0.0335}^{\ast\dagger\ddagger}$ & ${0.0115}^{\ast\dagger\ddagger}$ & ${0.0222}^{\ast\dagger\ddagger}$ \\
Square four-well & ${9.37{\times}10^{-4}}\,{}^{\ast\dagger\ddagger}$ & ${0.00174}^{\ast\dagger\ddagger}$ & ${0.00178}^{\ast\dagger\ddagger}$ & ${0.00235}^{\ast\dagger\ddagger}$ & ${0.00963}^{\ast\dagger\ddagger}$ \\
Triple-well Duffing & ${0.0708}^{\ast\dagger\ddagger}$ & ${0.0753}^{\ast\dagger\ddagger}$ & ${0.0825}^{\ast\dagger\ddagger}$ & ${0.0709}^{\ast\dagger\ddagger}$ & ${0.0751}^{\ast\dagger\ddagger}$ \\
SNIC multi-attractor & ${4.05{\times}10^{-4}}\,{}^{\ast\dagger\ddagger}$ & ${3.58{\times}10^{-4}}\,{}^{\ast\dagger\ddagger}$ & ${4.71{\times}10^{-4}}\,{}^{\ast\dagger\ddagger}$ & ${3.11{\times}10^{-4}}\,{}^{\ast\dagger\ddagger}$ & ${2.67{\times}10^{-4}}\,{}^{\ast\dagger\ddagger}$ \\
Transition-routes four-well & ${0.269}^{\ast\dagger\ddagger}$ & ${0.343}^{\ast\dagger\ddagger}$ & ${0.270}^{\ast\dagger\ddagger}$ & ${0.287}^{\ast\dagger\ddagger}$ & ${0.169}^{\ast\dagger\ddagger}$ \\
Depth-gradient four-well & ${8.65{\times}10^{-4}}\,{}^{\ast\dagger\ddagger}$ & ${0.00156}^{\ast\dagger\ddagger}$ & ${0.00943}^{\ast\dagger\ddagger}$ & ${0.00131}^{\ast\dagger\ddagger}$ & ${6.72{\times}10^{-4}}\,{}^{\ast\dagger\ddagger}$ \\
Diamond four-well & ${0.0797}^{\ast\dagger\ddagger}$ & ${0.0785}^{\ast\dagger\ddagger}$ & ${0.0894}^{\ast\dagger\ddagger}$ & ${0.114}^{\ast\dagger\ddagger}$ & ${0.0511}^{\ast\dagger\ddagger}$ \\
L-shaped five-well & ${0.00577}^{\ast\dagger\ddagger}$ & ${0.0254}^{\ast\dagger\ddagger}$ & ${0.0257}^{\ast\dagger\ddagger}$ & ${0.0167}^{\ast\dagger\ddagger}$ & ${5.04{\times}10^{-4}}\,{}^{\ast\dagger\ddagger}$ \\
\bottomrule
\end{tabular*}

\smallskip

\begin{tabular*}{\linewidth}{@{\extracolsep{\fill}}l cccccc@{}}
\toprule
 & & & \multicolumn{2}{c}{Poly.\ EDMD} & \multicolumn{2}{c}{RBF EDMD} \\
\cmidrule(lr){4-5}\cmidrule(l){6-7}
System & Dense MLP & DMD & direct & re-lifted & direct & re-lifted \\
\midrule
Local-linear gates & $0.236$ & $2.87$ & $2.80$ & $1.57$ & $3.22$ & $0.243$ \\
Transfer-gated local-linear & $1.21$ & $2.85$ & $3.82$ & $2.75$ & $2.71$ & $1.71$ \\
Arrested spiral & $2.69$ & $2.48$ & $2.46$ & $2.44$ & $2.87$ & $2.52$ \\
Asymmetric three-well & $3.15$ & $2.76$ & $2.76$ & $2.76$ & $3.09$ & $2.76$ \\
High-cross three-well & $5.01$ & $3.02$ & $3.00$ & $2.91$ & $3.07$ & $2.85$ \\
Hexagonal six-well & $4.82$ & $2.52$ & $2.52$ & $2.54$ & $7.74$ & $2.56$ \\
Octagonal eight-well & $8.18$ & $4.36$ & $4.37$ & $4.40$ & $6.42$ & $4.46$ \\
Pentagonal five-well & $4.92$ & $2.90$ & $2.90$ & $2.88$ & $5.75$ & $2.85$ \\
Square four-well & $0.293$ & $3.01$ & $3.02$ & $3.02$ & $4.94$ & $3.00$ \\
Triple-well Duffing & $0.431$ & $1.06$ & $1.06$ & $0.747$ & $1.03$ & $0.402$ \\
SNIC multi-attractor & $0.168$ & $0.810$ & $0.833$ & $0.854$ & $2.57$ & $0.192$ \\
Transition-routes four-well & $4.06$ & $5.26$ & $5.32$ & $5.32$ & $5.47$ & $5.26$ \\
Depth-gradient four-well & $0.247$ & $3.09$ & $3.09$ & $3.11$ & $3.64$ & $3.09$ \\
Diamond four-well & $1.33$ & $4.23$ & $4.23$ & $4.24$ & $4.67$ & $4.27$ \\
L-shaped five-well & $6.86$ & $3.31$ & $3.33$ & $3.30$ & $9.32$ & $3.37$ \\
\bottomrule
\end{tabular*}
}
\end{table}
\par

\subsection{Dysts benchmark}
\label{app:per_system_dysts}

\paragraph{Dense MLP.}
In \cref{tab:persystem_dysts_h100,tab:persystem_dysts_h2000,tab:persystem_dysts_h4000}, every sparse model has significantly lower error than Dense MLP in all \(10\) systems at every horizon, except Sparse MLP in Dequan Li at \(H=100\).

\paragraph{Classical baselines.}
The outcome depends on the system, and polynomial EDMD is a strong baseline on this benchmark: eight of the ten systems have quadratic vector fields (all but Chua and San-Um--Srisuchinwong), so its cubic dictionary contains their one-step map up to an error of third order in the step, which is about a thousandth of a period. As expected, re-lifted polynomial and RBF EDMD have significantly lower error than every sparse model in five of these systems, often by orders of magnitude (\cref{app:classical_baselines}). RBF EDMD, whose dictionary only approximates the one-step map, wins in the same systems by smaller margins.

\paragraph{Where the sparse models win.}
In the other five systems, all three LISTA models have significantly lower error than both re-lifted baselines at every horizon, and the two sparse MLPs do so in \(53\) of their \(60\) comparisons, never significantly higher. These five include three quadratic systems, Dadras, Sakarya, and Wang--Sun, where polynomial EDMD should have the advantage: its cubic dictionary contains the one-step map, whereas LISTA learns its features from data alone. Yet LISTA has significantly lower error in all three, and at \(H=100\) its error is about \(37\), \(3\), and \(22\) times lower than that of re-lifted polynomial EDMD, respectively. Here, a sparse KAE trained only on data outperforms a strong classical baseline whose dictionary matches the form of the dynamics.

\paragraph{System averages.}
\Cref{tab:forecasting_main} averages the per-system errors arithmetically, so each mean is the expected error on a randomly drawn system and emphasizes the hardest systems, where errors are largest.

\begin{table}[tbp]
\centering
\caption{Dysts test MSE per system at \(H=100\) observation steps, as in \cref{tab:persystem_mb_h100}.}
\label{tab:persystem_dysts_h100}
{\footnotesize\setlength{\tabcolsep}{3pt}\begin{tabular*}{\linewidth}{@{\extracolsep{\fill}}l ccccc@{}}
\toprule
System & LISTA & LISTA-BD & LISTA-SB & Sparse MLP-BD & Sparse MLP \\
\midrule
\addlinespace[1pt]
Chua & ${1.64{\times}10^{-5}}\,{}^{\ast\dagger\ddagger}$ & ${1.71{\times}10^{-5}}\,{}^{\ast\dagger\ddagger}$ & ${1.74{\times}10^{-5}}\,{}^{\ast\dagger\ddagger}$ & ${3.19{\times}10^{-5}}\,{}^{\ast\dagger\ddagger}$ & ${3.86{\times}10^{-5}}\,{}^{\ast\dagger\ddagger}$ \\
Dadras & ${1.70{\times}10^{-4}}\,{}^{\ast\dagger\ddagger}$ & ${1.78{\times}10^{-4}}\,{}^{\ast\dagger\ddagger}$ & ${1.74{\times}10^{-4}}\,{}^{\ast\dagger\ddagger}$ & ${4.24{\times}10^{-4}}\,{}^{\ast\dagger\ddagger}$ & ${4.40{\times}10^{-4}}\,{}^{\ast\dagger\ddagger}$ \\
Dequan Li & ${5.24{\times}10^{-5}}\,{}^{\ast\dagger}$ & ${5.15{\times}10^{-5}}\,{}^{\ast\dagger}$ & ${5.65{\times}10^{-5}}\,{}^{\ast\dagger}$ & ${1.10{\times}10^{-4}}\,{}^{\ast\dagger}$ & ${1.50{\times}10^{-4}}\,{}^{\dagger}$ \\
Hadley & ${4.68{\times}10^{-5}}\,{}^{\ast\dagger}$ & ${4.79{\times}10^{-5}}\,{}^{\ast\dagger}$ & ${4.50{\times}10^{-5}}\,{}^{\ast\dagger}$ & ${1.06{\times}10^{-4}}\,{}^{\ast\dagger}$ & ${1.34{\times}10^{-4}}\,{}^{\ast\dagger}$ \\
Lu--Chen--Cheng & ${7.83{\times}10^{-6}}\,{}^{\ast\dagger}$ & ${8.72{\times}10^{-6}}\,{}^{\ast\dagger}$ & ${8.27{\times}10^{-6}}\,{}^{\ast\dagger}$ & ${2.59{\times}10^{-5}}\,{}^{\ast\dagger}$ & ${2.99{\times}10^{-5}}\,{}^{\ast\dagger}$ \\
Qi--Chen & ${2.33{\times}10^{-5}}\,{}^{\ast\dagger}$ & ${2.73{\times}10^{-5}}\,{}^{\ast\dagger}$ & ${2.64{\times}10^{-5}}\,{}^{\ast\dagger}$ & ${7.43{\times}10^{-5}}\,{}^{\ast\dagger}$ & ${8.97{\times}10^{-5}}\,{}^{\ast\dagger}$ \\
Sakarya & ${1.82{\times}10^{-4}}\,{}^{\ast\dagger\ddagger}$ & ${2.24{\times}10^{-4}}\,{}^{\ast\dagger\ddagger}$ & ${2.42{\times}10^{-4}}\,{}^{\ast\dagger\ddagger}$ & ${7.26{\times}10^{-4}}\,{}^{\ast\dagger}$ & ${6.11{\times}10^{-4}}\,{}^{\ast\dagger}$ \\
San-Um--Srisuchinwong & ${3.79{\times}10^{-5}}\,{}^{\ast\dagger\ddagger}$ & ${4.15{\times}10^{-5}}\,{}^{\ast\dagger\ddagger}$ & ${3.98{\times}10^{-5}}\,{}^{\ast\dagger\ddagger}$ & ${1.02{\times}10^{-4}}\,{}^{\ast\dagger\ddagger}$ & ${1.11{\times}10^{-4}}\,{}^{\ast\dagger\ddagger}$ \\
Shimizu--Morioka & ${9.10{\times}10^{-6}}\,{}^{\ast\dagger}$ & ${9.19{\times}10^{-6}}\,{}^{\ast\dagger}$ & ${1.02{\times}10^{-5}}\,{}^{\ast\dagger}$ & ${2.70{\times}10^{-5}}\,{}^{\ast\dagger}$ & ${3.21{\times}10^{-5}}\,{}^{\ast\dagger}$ \\
Wang--Sun & ${2.14{\times}10^{-4}}\,{}^{\ast\dagger\ddagger}$ & ${2.59{\times}10^{-4}}\,{}^{\ast\dagger\ddagger}$ & ${2.27{\times}10^{-4}}\,{}^{\ast\dagger\ddagger}$ & ${5.26{\times}10^{-4}}\,{}^{\ast\dagger\ddagger}$ & ${6.14{\times}10^{-4}}\,{}^{\ast\dagger\ddagger}$ \\
\bottomrule
\end{tabular*}

\smallskip

\begin{tabular*}{\linewidth}{@{\extracolsep{\fill}}l cccccc@{}}
\toprule
 & & & \multicolumn{2}{c}{Poly.\ EDMD} & \multicolumn{2}{c}{RBF EDMD} \\
\cmidrule(lr){4-5}\cmidrule(l){6-7}
System & Dense MLP & DMD & direct & re-lifted & direct & re-lifted \\
\midrule
Chua & $1.02{\times}10^{-4}$ & $0.00521$ & $2.28{\times}10^{-4}$ & $1.36{\times}10^{-4}$ & $4.95{\times}10^{-5}$ & $4.69{\times}10^{-5}$ \\
Dadras & $0.0249$ & $1.39$ & $0.671$ & $0.00621$ & $0.0826$ & $0.0108$ \\
Dequan Li & $1.81{\times}10^{-4}$ & $0.576$ & $1.26{\times}10^{-4}$ & $3.76{\times}10^{-8}$ & $8.66{\times}10^{-6}$ & $1.35{\times}10^{-6}$ \\
Hadley & $2.57{\times}10^{-4}$ & $0.0831$ & $4.96{\times}10^{-5}$ & $1.09{\times}10^{-8}$ & $4.02{\times}10^{-6}$ & $3.64{\times}10^{-7}$ \\
Lu--Chen--Cheng & $7.05{\times}10^{-5}$ & $0.513$ & $8.73{\times}10^{-4}$ & $8.13{\times}10^{-8}$ & $1.39{\times}10^{-6}$ & $1.63{\times}10^{-7}$ \\
Qi--Chen & $8.74{\times}10^{-4}$ & $0.915$ & $0.0524$ & $8.49{\times}10^{-7}$ & $0.00111$ & $7.27{\times}10^{-6}$ \\
Sakarya & $0.0102$ & $1.13$ & $1.48$ & $6.07{\times}10^{-4}$ & $0.0156$ & $0.00458$ \\
San-Um--Srisuchinwong & $0.00190$ & $0.721$ & $0.0317$ & $0.00475$ & $6.27{\times}10^{-4}$ & $1.75{\times}10^{-4}$ \\
Shimizu--Morioka & $8.80{\times}10^{-5}$ & $0.715$ & $0.00555$ & $1.15{\times}10^{-10}$ & $9.39{\times}10^{-7}$ & $6.03{\times}10^{-7}$ \\
Wang--Sun & $0.0231$ & $1.42$ & $0.836$ & $0.00471$ & $0.0939$ & $0.00146$ \\
\bottomrule
\end{tabular*}
}
\end{table}

\begin{table}[tbp]
\centering
\caption{Dysts test MSE per system at \(H=2000\) observation steps, as in \cref{tab:persystem_mb_h100}.}
\label{tab:persystem_dysts_h2000}
{\footnotesize\setlength{\tabcolsep}{3pt}\begin{tabular*}{\linewidth}{@{\extracolsep{\fill}}l ccccc@{}}
\toprule
System & LISTA & LISTA-BD & LISTA-SB & Sparse MLP-BD & Sparse MLP \\
\midrule
\addlinespace[1pt]
Chua & ${0.00625}^{\ast\dagger\ddagger}$ & ${0.00738}^{\ast\dagger\ddagger}$ & ${0.0116}^{\ast\dagger\ddagger}$ & ${0.0181}^{\ast\dagger\ddagger}$ & ${0.0248}^{\ast\dagger\ddagger}$ \\
Dadras & ${0.777}^{\ast\dagger\ddagger}$ & ${0.803}^{\ast\dagger\ddagger}$ & ${0.927}^{\ast\dagger\ddagger}$ & ${1.16}^{\ast\dagger\ddagger}$ & ${1.26}^{\ast\dagger\ddagger}$ \\
Dequan Li & ${0.0505}^{\ast\dagger}$ & ${0.0446}^{\ast\dagger}$ & ${0.0559}^{\ast\dagger}$ & ${0.0781}^{\ast\dagger}$ & ${0.0826}^{\ast\dagger}$ \\
Hadley & ${0.0117}^{\ast\dagger}$ & ${0.0111}^{\ast\dagger}$ & ${0.0124}^{\ast\dagger}$ & ${0.0234}^{\ast\dagger}$ & ${0.0294}^{\ast\dagger}$ \\
Lu--Chen--Cheng & ${0.00140}^{\ast\dagger}$ & ${0.00176}^{\ast\dagger}$ & ${0.00170}^{\ast\dagger}$ & ${0.00490}^{\ast\dagger}$ & ${0.00629}^{\ast\dagger}$ \\
Qi--Chen & ${0.395}^{\ast\dagger}$ & ${0.329}^{\ast\dagger}$ & ${0.325}^{\ast\dagger}$ & ${0.456}^{\ast\dagger}$ & ${0.581}^{\ast\dagger}$ \\
Sakarya & ${0.499}^{\ast\dagger\ddagger}$ & ${0.491}^{\ast\dagger\ddagger}$ & ${0.543}^{\ast\dagger\ddagger}$ & ${0.670}^{\ast\dagger}$ & ${0.769}^{\ast\dagger}$ \\
San-Um--Srisuchinwong & ${0.270}^{\ast\dagger\ddagger}$ & ${0.300}^{\ast\dagger\ddagger}$ & ${0.354}^{\ast\dagger\ddagger}$ & ${0.399}^{\ast\dagger\ddagger}$ & ${0.481}^{\ast\dagger\ddagger}$ \\
Shimizu--Morioka & ${0.0894}^{\ast\dagger}$ & ${0.0847}^{\ast\dagger}$ & ${0.0863}^{\ast\dagger}$ & ${0.171}^{\ast\dagger}$ & ${0.170}^{\ast\dagger}$ \\
Wang--Sun & ${1.29}^{\ast\dagger\ddagger}$ & ${1.33}^{\ast\dagger\ddagger}$ & ${1.27}^{\ast\dagger\ddagger}$ & ${1.68}^{\ast\dagger\ddagger}$ & ${1.85}^{\ast\dagger\ddagger}$ \\
\bottomrule
\end{tabular*}

\smallskip

\begin{tabular*}{\linewidth}{@{\extracolsep{\fill}}l cccccc@{}}
\toprule
 & & & \multicolumn{2}{c}{Poly.\ EDMD} & \multicolumn{2}{c}{RBF EDMD} \\
\cmidrule(lr){4-5}\cmidrule(l){6-7}
System & Dense MLP & DMD & direct & re-lifted & direct & re-lifted \\
\midrule
Chua & $0.0627$ & $1.95$ & $0.899$ & $0.140$ & $0.725$ & $0.0768$ \\
Dadras & $2.00$ & $2.79$ & $2.72$ & $2.36$ & $2.40$ & $4.18$ \\
Dequan Li & $0.168$ & $2.82$ & $2.02$ & $3.27{\times}10^{-5}$ & $0.699$ & $0.00161$ \\
Hadley & $0.0783$ & $4.72$ & $2.84$ & $1.34{\times}10^{-5}$ & $1.15$ & $1.29{\times}10^{-4}$ \\
Lu--Chen--Cheng & $0.0173$ & $2.76$ & $1.12$ & $5.34{\times}10^{-5}$ & $0.592$ & $1.19{\times}10^{-4}$ \\
Qi--Chen & $0.783$ & $2.97$ & $2.81$ & $0.0587$ & $2.38$ & $0.175$ \\
Sakarya & $1.17$ & $2.77$ & $2.79$ & $0.861$ & $2.20$ & $1.98$ \\
San-Um--Srisuchinwong & $1.11$ & $3.15$ & $2.98$ & $3.83$ & $2.39$ & $0.605$ \\
Shimizu--Morioka & $0.245$ & $2.79$ & $2.48$ & $7.82{\times}10^{-7}$ & $1.71$ & $9.52{\times}10^{-4}$ \\
Wang--Sun & $2.88$ & $3.40$ & $3.37$ & $2.39$ & $3.03$ & $3.00$ \\
\bottomrule
\end{tabular*}
}
\end{table}

\begin{table}[tbp]
\centering
\caption{Dysts test MSE per system at \(H=4000\) observation steps, as in \cref{tab:persystem_mb_h100}.}
\label{tab:persystem_dysts_h4000}
{\footnotesize\setlength{\tabcolsep}{3pt}\begin{tabular*}{\linewidth}{@{\extracolsep{\fill}}l ccccc@{}}
\toprule
System & LISTA & LISTA-BD & LISTA-SB & Sparse MLP-BD & Sparse MLP \\
\midrule
\addlinespace[1pt]
Chua & ${0.117}^{\ast\dagger\ddagger}$ & ${0.105}^{\ast\dagger\ddagger}$ & ${0.129}^{\ast\dagger\ddagger}$ & ${0.209}^{\ast\dagger\ddagger}$ & ${0.274}^{\ast\dagger\ddagger}$ \\
Dadras & ${2.50}^{\ast\dagger\ddagger}$ & ${2.54}^{\ast\dagger\ddagger}$ & ${2.58}^{\ast}$ & ${2.92}^{\ast}$ & ${3.03}^{\ast}$ \\
Dequan Li & ${0.127}^{\ast\dagger}$ & ${0.126}^{\ast\dagger}$ & ${0.148}^{\ast\dagger}$ & ${0.189}^{\ast\dagger}$ & ${0.233}^{\ast\dagger}$ \\
Hadley & ${0.108}^{\ast\dagger}$ & ${0.102}^{\ast\dagger}$ & ${0.121}^{\ast\dagger}$ & ${0.179}^{\ast\dagger}$ & ${0.216}^{\ast\dagger}$ \\
Lu--Chen--Cheng & ${0.00822}^{\ast\dagger}$ & ${0.00999}^{\ast\dagger}$ & ${0.00988}^{\ast\dagger}$ & ${0.0321}^{\ast\dagger}$ & ${0.0355}^{\ast\dagger}$ \\
Qi--Chen & ${2.03}^{\ast\dagger}$ & ${1.92}^{\ast\dagger}$ & ${1.95}^{\ast\dagger}$ & ${2.31}^{\ast\dagger}$ & ${2.46}^{\ast\dagger}$ \\
Sakarya & ${1.75}^{\ast\dagger\ddagger}$ & ${1.79}^{\ast\dagger\ddagger}$ & ${1.80}^{\ast\dagger\ddagger}$ & ${2.23}^{\ast\dagger\ddagger}$ & ${2.33}^{\ast\dagger\ddagger}$ \\
San-Um--Srisuchinwong & ${2.07}^{\ast\dagger\ddagger}$ & ${2.23}^{\ast\dagger\ddagger}$ & ${2.25}^{\ast\dagger\ddagger}$ & ${2.51}^{\ast\dagger\ddagger}$ & ${2.61}^{\ast\dagger\ddagger}$ \\
Shimizu--Morioka & ${0.361}^{\ast\dagger}$ & ${0.311}^{\ast\dagger}$ & ${0.395}^{\ast\dagger}$ & ${0.687}^{\ast\dagger}$ & ${0.676}^{\ast\dagger}$ \\
Wang--Sun & ${3.66}^{\ast}$ & ${3.73}^{\ast}$ & ${3.65}^{\ast}$ & ${4.02}^{\ast}$ & ${4.15}^{\ast}$ \\
\bottomrule
\end{tabular*}

\smallskip

\begin{tabular*}{\linewidth}{@{\extracolsep{\fill}}l cccccc@{}}
\toprule
 & & & \multicolumn{2}{c}{Poly.\ EDMD} & \multicolumn{2}{c}{RBF EDMD} \\
\cmidrule(lr){4-5}\cmidrule(l){6-7}
System & Dense MLP & DMD & direct & re-lifted & direct & re-lifted \\
\midrule
Chua & $0.518$ & $3.52$ & $1.66$ & $0.958$ & $1.51$ & $0.344$ \\
Dadras & $3.64$ & $2.83$ & $2.79$ & $3.94$ & $2.63$ & $6.56$ \\
Dequan Li & $0.388$ & $2.75$ & $2.84$ & $3.44{\times}10^{-4}$ & $1.10$ & $0.0152$ \\
Hadley & $0.399$ & $4.70$ & $4.23$ & $1.30{\times}10^{-4}$ & $2.03$ & $0.00121$ \\
Lu--Chen--Cheng & $0.0819$ & $2.88$ & $1.67$ & $6.96{\times}10^{-4}$ & $1.11$ & $9.22{\times}10^{-4}$ \\
Qi--Chen & $2.94$ & $3.05$ & $2.96$ & $0.917$ & $2.75$ & $1.38$ \\
Sakarya & $2.79$ & $2.80$ & $2.81$ & $2.90$ & $2.50$ & $3.82$ \\
San-Um--Srisuchinwong & $3.36$ & $3.27$ & $3.19$ & $5.22$ & $2.90$ & $2.88$ \\
Shimizu--Morioka & $0.929$ & $2.86$ & $2.70$ & $1.87{\times}10^{-5}$ & $2.25$ & $0.0407$ \\
Wang--Sun & $4.70$ & $3.45$ & $3.44$ & $4.59$ & $3.27$ & $4.91$ \\
\bottomrule
\end{tabular*}
}
\end{table}
\par

\FloatBarrier\section{Multibasin benchmark inventory and system definitions}
\label{app:multibasin_inventory}

\Cref{tab:fixed15_inventory} lists the \(15\) two-dimensional multibasin systems \(\dot x=f_s(x)\) used in this paper, and \cref{fig:benchmarks} shows four of them. Claude Opus 4.5 in Claude Code provided substantial help in procedurally generating candidates given initial specifications, from which these 15 systems were chosen.

\begin{table}[tbp]
\centering
\caption{The \(15\) multibasin benchmark systems. Every system has state dimension \(d=2\). Each stored observation is one fourth-order Runge--Kutta step of size \(\Delta t\) of \(\dot x=f_s(x)\) from the previous stored state, and initial conditions are drawn uniformly from the listed box. Parameters of the Gaussian-well systems are in \cref{tab:app_gaussian_params}; the other systems are defined in the text. ``Basins'' is the number of designed attractors and of trajectory-fate labels (\cref{app:exp_alignment}). On Transfer-gated local-linear, about \(40\%\) of uniform starts instead stall at region boundaries, and the labeling groups these stall points with the designed attractors.}
\label{tab:fixed15_inventory}
{\footnotesize
\setlength{\tabcolsep}{3pt}
\begin{tabular}{@{}>{\raggedright\arraybackslash}p{3.1cm} >{\raggedright\arraybackslash}p{2.4cm} >{\raggedright\arraybackslash}p{3.9cm} c c >{\centering\arraybackslash}p{1.8cm} c@{}}
\toprule
Display name and system key & Generator family & Parameters & \(d\) & \(\Delta t\) & Initial-condition box & Basins \\
\midrule
Local-linear gates \newline {\scriptsize\texttt{gated\_local\_linear}}
  & piecewise local-linear gates
  & \(c_b=1.75(\cos\alpha_b,\sin\alpha_b)\), core radius \(1.05\); \(\widetilde A_b\), \(G\) in text
  & 2 & 0.04 & \([-2.6,2.6]^2\) & 3 \\
Transfer-gated local-linear \newline {\scriptsize\texttt{gated\_transfer\_linear}}
  & piecewise transfer gates
  & \(c_b=1.85(\cos\alpha_b,\sin\alpha_b)\); radii, widths, speeds, and templates in text
  & 2 & 0.04 & \([-2.8,2.8]^2\) & 3 \\
Arrested spiral \newline {\scriptsize\texttt{arrested\_spiral}}
  & spiral capture with Gaussian traps
  & damping \(0.3\), rotation \(2.0\); traps at \((1.5,0)\), \((0,1.8)\), \((-1,-1)\), \((0.8,-1.5)\), and the origin
  & 2 & 0.02 & \([-3,3]^2\) & 5 \\
Asymmetric three-well \newline {\scriptsize\texttt{cal\_asymmetric\_3}}
  & Gaussian wells with independent rotation
  & \cref{tab:app_gaussian_params}
  & 2 & 0.03 & \([-3,3]\times[-2.5,3]\) & 3 \\
High-cross three-well \newline {\scriptsize\texttt{cal\_high\_cross\_3}}
  & Gaussian wells with stronger rotation
  & \cref{tab:app_gaussian_params}
  & 2 & 0.03 & \([-3.3,3.3]^2\) & 3 \\
Hexagonal six-well \newline {\scriptsize\texttt{cal\_hexagon\_6}}
  & Gaussian wells with independent rotation
  & \cref{tab:app_gaussian_params}
  & 2 & 0.03 & \([-3.3,3.3]^2\) & 6 \\
Octagonal eight-well \newline {\scriptsize\texttt{cal\_octagon\_8}}
  & Gaussian wells with independent rotation
  & \cref{tab:app_gaussian_params}
  & 2 & 0.03 & \([-4,4]^2\) & 8 \\
Pentagonal five-well \newline {\scriptsize\texttt{cal\_pentagon\_5}}
  & Gaussian wells with independent rotation
  & \cref{tab:app_gaussian_params}
  & 2 & 0.03 & \([-3.3,3.3]^2\) & 5 \\
Square four-well \newline {\scriptsize\texttt{cal\_square\_4}}
  & Gaussian wells with independent rotation
  & \cref{tab:app_gaussian_params}
  & 2 & 0.03 & \([-3.3,3.3]^2\) & 4 \\
Triple-well Duffing \newline {\scriptsize\texttt{duffing\_triple\_well\_}\allowbreak\texttt{no\_rotation}}
  & triple-well Duffing oscillator without rotation
  & \(a=0.3\), \(\delta=0.5\), \(\omega=0\), \(\epsilon=0.003\), \(s=4.0\) (\cref{eq:app_duffing_triple})
  & 2 & 0.02 & \([-2,2]^2\) & 3 \\
SNIC multi-attractor \newline {\scriptsize\texttt{snic\_multi}}
  & polar SNIC-like phase dynamics
  & \(\dot\theta=1-1.2\cos3\theta\), radial forcing \(0.3\), rotation \(0.5\), confinement \(0.01\)
  & 2 & 0.02 & \([-2,2]^2\) & 3 \\
Transition-routes four-well \newline {\scriptsize\texttt{transition\_routes\_4}}
  & route-augmented Gaussian wells
  & centers \((\pm1.8,\pm1.8)\), \((a,\sigma)=(3.0,0.6)\), \((\omega,\gamma)=(1.0,0.03)\), route gain \(0.3\) (\cref{eq:app_transition_routes})
  & 2 & 0.02 & \([-3.5,3.5]^2\) & 4 \\
Depth-gradient four-well \newline {\scriptsize\texttt{var\_depth\_gradient\_4}}
  & Gaussian wells with depth gradient
  & \cref{tab:app_gaussian_params}
  & 2 & 0.03 & \([-3,3]^2\) & 4 \\
Diamond four-well \newline {\scriptsize\texttt{var\_diamond\_4}}
  & Gaussian wells on a diamond layout
  & \cref{tab:app_gaussian_params}
  & 2 & 0.03 & \([-3.5,3.5]^2\) & 4 \\
L-shaped five-well \newline {\scriptsize\texttt{var\_l\_shape\_5}}
  & Gaussian wells on an L-shaped layout
  & \cref{tab:app_gaussian_params}
  & 2 & 0.03 & \([-3.5,3.5]^2\) & 5 \\
\bottomrule
\end{tabular}}
\end{table}

\begin{figure}[p]
\centering
\includegraphics[width=\textwidth]{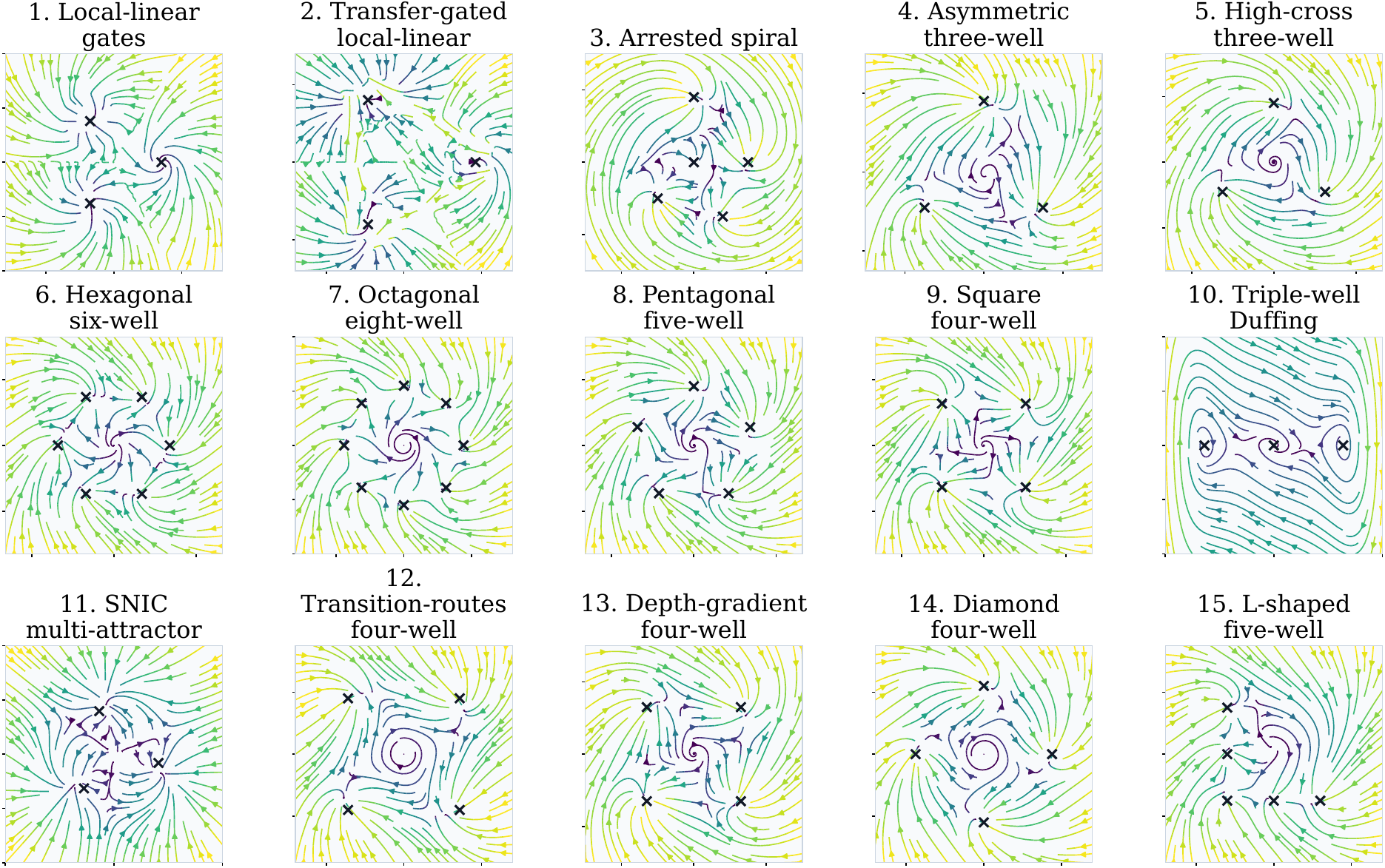}
\caption{Ground-truth vector fields for the retained \(15\)-system multibasin
benchmark. The models are trained only on stored state trajectories; these
continuous-time vector fields and design-center annotations are shown here only to
document the benchmark geometry and are not provided to the learner.}
\label{fig:app_ground_truth_vector_fields_retained15}
\end{figure}

\paragraph{Ground-truth vector-field visualizations.}
\Cref{fig:app_ground_truth_vector_fields_retained15} shows the continuous-time
ground-truth vector fields for all \(15\) retained two-dimensional multibasin
systems. Streamlines show the direction of \(f_s(x)\), and black crosses mark
the generator design centers (well, trap, or gate centers; for SNIC multi-attractor, the equilibria of its polar part without the added rotation and confinement terms), which can lie up to about \(0.5\) from the attractors; evaluation labels come from trajectory fates (\cref{app:exp_alignment}). The streamline colors use a panel-local
\(\log_{10}\norm{f_s(x)}_2\) scale to reveal structure within each system; they
should not be read as comparable speed scales across panels.
\begingroup\def\cref#1{Figures 7 to 11}\endgroup\par

\par
\paragraph{Gaussian-well family.}
Most systems use a common two-dimensional Gaussian potential
with an independent rotational drift. For \(x=(x_1,x_2)\in\R^2\), let
\(R x=(x_2,-x_1)\) and
\[
  V_s(x)
  =
  \sum_{i=1}^{M_s}
  -a_{s,i}
  \exp\!\left(-\frac{\norm{x-c_{s,i}}_2^2}{2\sigma_{s,i}^2}\right)
  +\gamma_s(x_1^4+x_2^4).
\]
The implemented vector field is
\begin{equation}
  \dot x = -\nabla V_s(x)+\omega_s R x .
  \label{eq:app_gaussian_well}
\end{equation}
Define \(p_i^{(M)}(r,\varphi)
=r(\cos(2\pi i/M+\varphi),\sin(2\pi i/M+\varphi))\), \(i=0,\ldots,M-1\).
The retained Gaussian-well systems instantiate \cref{eq:app_gaussian_well} with
the parameters in \cref{tab:app_gaussian_params}.

\begin{table}[tbp]
\centering
\caption{Parameters for retained Gaussian-well systems. Repeated pairs in the
\((a_i,\sigma_i)\) column apply to every listed center.}
\label{tab:app_gaussian_params}
\resizebox{\textwidth}{!}{\begin{tabular}{@{}l l l l@{}}
\toprule
System & \(\{c_i\}\) & \((a_i,\sigma_i)\) & \((\omega,\gamma)\) \\
\midrule
High-cross three-well
  & \(\{p_i^{(3)}(1.8,\pi/2)\}\)
  & \((3.0,0.5)\) & \((2.0,0.03)\) \\
Hexagonal six-well
  & \(\{p_i^{(6)}(1.7,0)\}\)
  & \((2.0,0.5)\) & \((1.0,0.03)\) \\
Octagonal eight-well
  & \(\{p_i^{(8)}(2.2,0)\}\)
  & \((3.0,0.5)\) & \((0.90,0.02)\) \\
Pentagonal five-well
  & \(\{p_i^{(5)}(1.8,\pi/2)\}\)
  & \((2.0,0.5)\) & \((1.1,0.03)\) \\
Square four-well
  & \(\{p_i^{(4)}(1.8,\pi/4)\}\)
  & \((3.0,0.5)\) & \((1.0,0.03)\) \\
Diamond four-well
  & \(\{(0,2.2),(2.2,0),(0,-2.2),(-2.2,0)\}\)
  & \((2.5,0.5)\) & \((1.0,0.02)\) \\
L-shaped five-well
  & \(\{(-1.5,1.5),(-1.5,0),(-1.5,-1.5),(0,-1.5),(1.5,-1.5)\}\)
  & \((2.5,0.5)\) & \((1.0,0.03)\) \\
Asymmetric three-well
  & \(\{(0,1.8),(-1.5,-0.9),(1.5,-0.9)\}\)
  & \(\{(2.5,0.55),(1.5,0.4),(2.0,0.5)\}\) & \((1.0,0.03)\) \\
Depth-gradient four-well
  & \(\{(-1.3,1.3),(1.3,1.3),(1.3,-1.3),(-1.3,-1.3)\}\)
  & \(\{(2.2,0.55),(2.5,0.5),(3.0,0.5),(3.5,0.5)\}\) & \((1.3,0.03)\) \\
\bottomrule
\end{tabular}
}
\end{table}

The transition-routes system uses the same Gaussian-well form with centers
\[
  \{c_i\}
  =
  \{(-1.8,1.8),(1.8,1.8),(-1.8,-1.8),(1.8,-1.8)\},
\]
\(a_i=3.0\), \(\sigma_i=0.6\), \(\omega=1.0\), and \(\gamma=0.03\), and adds a
corridor-dependent rotation term:
\begin{equation}
  \dot x
  =
  -\nabla V_s(x)
  +
  \bigl(1.0+0.3\,C_{\rm route}(x_2)\bigr) R x,
  \qquad
  C_{\rm route}(x_2)
  =
  e^{-(x_2-1.8)^2/0.3}+e^{-(x_2+1.8)^2/0.3}.
  \label{eq:app_transition_routes}
\end{equation}

\paragraph{Native gated local-linear systems.}
Both native gated systems place three centers on a circular layout, but with
different radii and region definitions. Let
\(Q(\alpha)=\begin{psmallmatrix}\cos\alpha&-\sin\alpha\\
\sin\alpha&\cos\alpha\end{psmallmatrix}\) and
\(\alpha_b=2\pi b/3\), \(b\in\{0,1,2\}\).

For \texttt{gated\_local\_linear}, \(c_b=1.75(\cos\alpha_b,\sin\alpha_b)\).
Inside the radius-\(1.05\) core of basin \(b\),
\[
  \dot x=A_b(x-c_b),\qquad
  A_b=Q(\alpha_b)\widetilde A_b Q(\alpha_b)^\top,
\]
with
\[
\widetilde A_0=\begin{psmallmatrix}-0.9&-1.2\\1.2&-0.9\end{psmallmatrix},
\quad
\widetilde A_1=\begin{psmallmatrix}-1.35&0.2\\-0.3&-0.7\end{psmallmatrix},
\quad
\widetilde A_2=\begin{psmallmatrix}-0.7&-0.1\\0.5&-1.2\end{psmallmatrix}.
\]
Outside the cores, the active sector is the nearest angular sector \(s(x)\) and
the shared gate matrix
\[
  G=\begin{psmallmatrix}-1.35&-0.9\\0.9&-1.35\end{psmallmatrix}
\]
drives the state as \(\dot x=G(x-c_{s(x)})\).

For \texttt{gated\_transfer\_linear}, \(c_b=1.85(\cos\alpha_b,\sin\alpha_b)\).
The implemented radii and widths are
\[
\begin{aligned}
  \rho_{\rm core}&=0.30, &
  \rho_{\rm source}&=0.80, &
  \rho_{\rm exit}&=0.60, &
  \beta_{\rm exit}&=0.72,\\
  w_{\rm chan}&=0.22, &
  \delta_{\rm lane}&=0.28, &
  \rho_{\rm hand}&=0.45 .
\end{aligned}
\]
The core matrices use the same rotated form with templates
\[
\begin{psmallmatrix}-1.0&-1.1\\1.1&-1.0\end{psmallmatrix},\quad
\begin{psmallmatrix}-1.4&0.2\\-0.2&-0.7\end{psmallmatrix},\quad
\begin{psmallmatrix}-0.8&-0.3\\0.5&-1.3\end{psmallmatrix}.
\]
For each ordered pair \(p=(s,t)\), \(s\ne t\), define
\[
  d_{s\to t}=\frac{c_t-c_s}{\norm{c_t-c_s}_2},\qquad
  n_{s\to t}=(-d_{s\to t,2},d_{s\to t,1}),\qquad
  o_t=\frac{c_t}{\norm{c_t}_2},
\]
and \(\chi_{s,t}=1\) when \(\sin(\alpha_s-\alpha_t)\ge 0\), otherwise
\(\chi_{s,t}=-1\). The implemented channel entry and handoff points are
\[
\begin{aligned}
  e_p &= c_s+\rho_{\rm source}d_{s\to t}
        +\chi_{s,t}\delta_{\rm lane}n_{s\to t},\\
  h_p &= c_t+\rho_{\rm hand}o_t
        +\chi_{s,t}\delta_{\rm lane}(-o_{t,2},o_{t,1}),
\end{aligned}
\]
with channel direction \(d_p=(h_p-e_p)/\norm{h_p-e_p}_2\), channel normal
\(n_p=(-d_{p,2},d_{p,1})\), and channel length
\(\ell_p=(h_p-e_p)\cdot d_p\).

The vector field is piecewise analytic. Core regions
\(\norm{x-c_b}_2\le\rho_{\rm core}\) use \(A_b(x-c_b)\). Source annuli use
\(0.65A_b(x-c_b)\), and other non-channel background regions use
\(0.50A_b(x-c_b)\) for the nearest center \(c_b\). Exit wedges for
\(p=(s,t)\) are source-neighborhood states outside the core with
\(\norm{x-c_s}_2\ge\rho_{\rm exit}\) and
\((x-c_s)\cdot d_{s\to t}\ge\norm{x-c_s}_2\cos\beta_{\rm exit}\); they use
\[
  \dot x = v_{\rm exit}d_{s\to t}
  -\lambda_{\rm exit}\bigl((x-c_s)\cdot n_{s\to t}\bigr)n_{s\to t},
  \qquad
  v_{\rm exit}=1.0,\quad \lambda_{\rm exit}=1.8;
\]
and channel rectangles satisfying
\[
  0\le (x-e_p)\cdot d_p\le \ell_p,\qquad
  |(x-e_p)\cdot n_p|\le w_{\rm chan}
\]
use
\[
  \dot x = v_{\rm chan}d_p
  -\lambda_{\rm chan}\bigl((x-e_p)\cdot n_p\bigr)n_p,
  \qquad
  v_{\rm chan}=1.55,\quad \lambda_{\rm chan}=2.8.
\]
Channel rectangles take precedence over all other regions, including the cores.

\paragraph{Other specialized systems.}
The arrested spiral system has a background spiral flow plus four Gaussian
traps and an origin trap:
\[
  \dot x
  =
  -0.3x+2.0Rx
  -4.0\sum_{i=1}^{4}
  e^{-\norm{x-c_i}_2^2/(2\cdot0.25)}
  \frac{x-c_i}{0.25}
  -1.5e^{-\norm{x}_2^2/0.3}\frac{x}{0.15}
  -0.02(x_1^3,x_2^3),
\]
with \(c_i\in\{(1.5,0),(0,1.8),(-1,-1),(0.8,-1.5)\}\). Its fifth basin is the
origin trap.

The triple-well Duffing system uses \(x=(q,p)\) and
\[
  V(q)=q^6/6-q^4/2+a q^2,\qquad
  V'(q)=q^5-2q^3+2a q,
\]
with \(a=0.3\), damping \(\delta=0.5\), rotation strength \(\omega=0\) (no rotation, so that each of the three wells holds an attractor), and
soft cubic confinement \(-\epsilon u^3/s^3\) with \(\epsilon=0.003\) and
\(s=4.0\):
\begin{equation}
\begin{aligned}
  \dot q &= (1+\omega)p-\epsilon q^3/s^3,\\
  \dot p &= -V'(q)-\delta p-\omega q-\epsilon p^3/s^3 .
\end{aligned}
\label{eq:app_duffing_triple}
\end{equation}

The SNIC system is implemented in polar form and then converted to Cartesian
coordinates. With \(\varepsilon_{\rm num}=10^{-8}\),
\(r^2=x_1^2+x_2^2+\varepsilon_{\rm num}\), \(r=\sqrt{r^2}\),
\(\theta=\operatorname{atan2}(x_2,x_1)\),
\(c_\theta=x_1/(r+\varepsilon_{\rm num})\), and
\(s_\theta=x_2/(r+\varepsilon_{\rm num})\),
\[
  \dot r=r(1-r^2)-0.3\cos(3\theta),\qquad
  \dot\theta=1-1.2\cos(3\theta).
\]
The Cartesian vector field is
\[
\begin{aligned}
  \dot x_1
  &= \dot r c_\theta-r\dot\theta s_\theta
     -0.01x_1(x_1^2+x_2^2)+0.5x_2,\\
  \dot x_2
  &= \dot r s_\theta+r\dot\theta c_\theta
     -0.01x_2(x_1^2+x_2^2)-0.5x_1 .
\end{aligned}
\]

\FloatBarrier\section{Dysts benchmark inventory and system definitions}
\label{app:dysts_inventory}

\Cref{tab:dysts_inventory_full} lists the ten Dysts systems \citep{gilpin_chaos_2021,gilpin_model_2023} of the forecasting benchmark, with right-hand sides and metadata from \texttt{dysts} \(0.96\) (Apache-2.0 license). The equations below are in native coordinates \((x,y,z)\); trajectories are integrated and stored in Dysts' standardized coordinates (per-coordinate mean and standard deviation from the Dysts metadata). All right-hand sides are closed-form; most are polynomial, and the Chua and San-Um--Srisuchinwong systems include absolute values, so they are only piecewise smooth.
\par
\begin{table}[tbp]
\centering
\caption{The ten Dysts systems used in the forecasting
benchmark. The ``stored step'' column is \(5\) times the native Dysts
integration interval, which gives \(999.8\) stored steps per dominant period for every system.}
\label{tab:dysts_inventory_full}
\resizebox{\textwidth}{!}{\begin{tabular}{@{}l l c c c@{}}
\toprule
Display name & System key & Dimension & Native \(dt\) & Stored step \(5dt\) \\
\midrule
Chua & \texttt{Chua} & 3 & \(2.8474745791{\times}10^{-4}\) & \(1.4237372895{\times}10^{-3}\) \\
Dadras & \texttt{Dadras} & 3 & \(6.5782963827{\times}10^{-4}\) & \(3.2891481914{\times}10^{-3}\) \\
Dequan Li & \texttt{DequanLi} & 3 & \(1.6763993999{\times}10^{-5}\) & \(8.3819969995{\times}10^{-5}\) \\
Hadley & \texttt{Hadley} & 3 & \(2.9086847808{\times}10^{-4}\) & \(1.4543423904{\times}10^{-3}\) \\
Lu--Chen--Cheng & \texttt{LuChenCheng} & 3 & \(1.8469678280{\times}10^{-4}\) & \(9.2348391399{\times}10^{-4}\) \\
Qi--Chen & \texttt{QiChen} & 3 & \(7.8371061844{\times}10^{-5}\) & \(3.9185530922{\times}10^{-4}\) \\
Sakarya & \texttt{Sakarya} & 3 & \(9.9704617436{\times}10^{-4}\) & \(4.9852308718{\times}10^{-3}\) \\
San-Um--Srisuchinwong & \texttt{SanUmSrisuchinwong} & 3 & \(1.4933288881{\times}10^{-3}\) & \(7.4666444407{\times}10^{-3}\) \\
Shimizu--Morioka & \texttt{ShimizuMorioka} & 3 & \(2.4080013336{\times}10^{-3}\) & \(1.2040006668{\times}10^{-2}\) \\
Wang--Sun & \texttt{WangSun} & 3 & \(5.3924987498{\times}10^{-3}\) & \(2.6962493749{\times}10^{-2}\) \\
\bottomrule
\end{tabular}
}
\end{table}

\paragraph{Trajectories and splits.}
Trajectories are integrated with DOP853 (relative and absolute tolerances \(10^{-12}\)). Each split holds \(200\) trajectories per system of \(30{,}001\) stored states, after a discarded warm-up of \(2{,}000\) stored steps. The first training trajectory starts from the Dysts default initial condition; every other trajectory perturbs it with Gaussian noise of \(0.2\) times the coordinate-wise Dysts standard deviation, drawn from a separate deterministic seed stream per split, so no validation or test trajectory is a training trajectory.

\paragraph{Sakarya trajectories.}
Sakarya's flow is unbounded along the \(z\)-axis, so a Sakarya trajectory whose standardized state norm exceeds ten times the largest one in the training split is replaced by the next initial condition of its split's stream. This replaced four of the \(600\) Sakarya trajectories, two in validation and two in test.
\par
\paragraph{Chua.}
The Chua system uses the piecewise-linear diode nonlinearity
\[
  h(x)
  =
  m_1x+\frac{1}{2}(m_0-m_1)\bigl(|x+1|-|x-1|\bigr),
\]
with parameters
\[
  \alpha=15.6,\qquad
  \beta=28,\qquad
  m_0=-1.142857,\qquad
  m_1=-0.71429 .
\]
The vector field is
\begin{equation}
\begin{aligned}
  \dot x &= \alpha\bigl(y-x-h(x)\bigr),\\
  \dot y &= x-y+z,\\
  \dot z &= -\beta y .
\end{aligned}
\label{eq:app_dysts_chua}
\end{equation}

\paragraph{Dadras.}
With parameters
\[
  c=2,\qquad e=9,\qquad o=2.7,\qquad p=3,\qquad r=1.7,
\]
the Dadras system is
\begin{equation}
\begin{aligned}
  \dot x &= y-px+oyz,\\
  \dot y &= ry-xz+z,\\
  \dot z &= cxy-ez .
\end{aligned}
\label{eq:app_dysts_dadras}
\end{equation}

\paragraph{Dequan Li.}
With parameters
\[
  a=40,\qquad c=1.833,\qquad d=0.16,\qquad
  \varepsilon=0.65,\qquad f=20,\qquad k=55,
\]
the Dequan Li system is
\begin{equation}
\begin{aligned}
  \dot x &= a(y-x)+dxz,\\
  \dot y &= kx+fy-xz,\\
  \dot z &= cz+xy-\varepsilon x^2 .
\end{aligned}
\label{eq:app_dysts_dequan_li}
\end{equation}

\paragraph{Hadley.}
With parameters
\[
  a=0.2,\qquad b=4,\qquad f=9,\qquad g=1,
\]
the Hadley system is
\begin{equation}
\begin{aligned}
  \dot x &= -y^2-z^2-ax+af,\\
  \dot y &= xy-bxz-y+g,\\
  \dot z &= bxy+xz-z .
\end{aligned}
\label{eq:app_dysts_hadley}
\end{equation}

\paragraph{Lu--Chen--Cheng.}
With parameters
\[
  a=-10,\qquad b=-4,\qquad c=18.1,
\]
the Lu--Chen--Cheng system is
\begin{equation}
\begin{aligned}
  \dot x &= -\frac{ab}{a+b}x-yz+c,\\
  \dot y &= ay+xz,\\
  \dot z &= bz+xy .
\end{aligned}
\label{eq:app_dysts_lu_chen_cheng}
\end{equation}

\paragraph{Qi--Chen.}
With parameters
\[
  a=38,\qquad b=2.666,\qquad c=80,
\]
the Qi--Chen system is
\begin{equation}
\begin{aligned}
  \dot x &= a(y-x)+yz,\\
  \dot y &= cx+y-xz,\\
  \dot z &= xy-bz .
\end{aligned}
\label{eq:app_dysts_qi_chen}
\end{equation}

\paragraph{Sakarya.}
With parameters
\[
  a=-1,\qquad b=1,\qquad c=1,\qquad h=1,\qquad
  p=1,\qquad q=0.4,\qquad r=0.3,\qquad s=1,
\]
the Sakarya system is
\begin{equation}
\begin{aligned}
  \dot x &= ax+hy+syz,\\
  \dot y &= -by-px+qxz,\\
  \dot z &= cz-rxy .
\end{aligned}
\label{eq:app_dysts_sakarya}
\end{equation}

\paragraph{San-Um--Srisuchinwong.}
With parameter \(a=2\), the San-Um--Srisuchinwong system is
\begin{equation}
\begin{aligned}
  \dot x &= y-x,\\
  \dot y &= -z\tanh(x),\\
  \dot z &= -a+xy+|y| .
\end{aligned}
\label{eq:app_dysts_san_um_srisuchinwong}
\end{equation}

\paragraph{Shimizu--Morioka.}
With parameters
\[
  a=0.85,\qquad b=0.5,
\]
the Shimizu--Morioka system is
\begin{equation}
\begin{aligned}
  \dot x &= y,\\
  \dot y &= x-ay-xz,\\
  \dot z &= -bz+x^2 .
\end{aligned}
\label{eq:app_dysts_shimizu_morioka}
\end{equation}

\paragraph{Wang--Sun.}
With parameters
\[
  a=0.2,\qquad b=-0.01,\qquad d=-0.4,\qquad
  e=-1,\qquad f=-1,\qquad q=1,
\]
the Wang--Sun system is
\begin{equation}
\begin{aligned}
  \dot x &= ax+qyz,\\
  \dot y &= bx+dy-xz,\\
  \dot z &= ez+fxy .
\end{aligned}
\label{eq:app_dysts_wang_sun}
\end{equation}

\FloatBarrier\par

% \newpage
% \input{checklist.tex}

\end{document}